\documentclass[pmlr,twocolumn,10pt]{jmlr} 

\usepackage[utf8]{inputenc}
\usepackage{xcolor}
\usepackage{amsfonts}
\usepackage{booktabs}
\usepackage{multirow}
\usepackage{adjustbox}
\usepackage[normalem]{ulem}
\usepackage{enumitem}
\useunder{\uline}{\ul}{}
\usepackage{todonotes}
\usepackage{algorithm,algorithmic}

\jmlrvolume{LEAVE UNSET}
\jmlryear{2022}
\jmlrsubmitted{LEAVE UNSET}
\jmlrpublished{LEAVE UNSET}
\jmlrworkshop{Conference on Health, Inference, and Learning (CHIL) 2022} 

\title{Data Augmentation for Electrocardiograms}

\author{%
\Name{Aniruddh Raghu} \Email{araghu@mit.edu}\\
\addr Massachusetts Institute of Technology, USA
\AND
\Name{Divya Shanmugam} \Email{divyas@mit.edu}\\
\addr Massachusetts Institute of Technology, USA
\AND
\Name{Eugene Pomerantsev} \Email{epomerantsev@mgh.harvard.edu}\\
\addr Massachusetts General Hospital, USA
\AND
\Name{John Guttag} \Email{guttag@mit.edu}\\
\addr Massachusetts Institute of Technology, USA
\AND
\Name{Collin M. Stultz} \Email{cmstultz@mit.edu}\\
\addr Massachusetts Institute of Technology, USA
}

\begin{document}
\maketitle
\begin{abstract}
Neural network models have demonstrated impressive performance in predicting pathologies and outcomes from the 12-lead electrocardiogram (ECG). However, these models often need to be trained with large, labelled datasets, which are not available for many predictive tasks of interest.
In this work, we perform an empirical study examining whether training time data augmentation methods can be used to improve performance on such data-scarce ECG prediction problems. 
We investigate how data augmentation strategies impact model performance when detecting cardiac abnormalities from the ECG.
Motivated by our finding that the effectiveness of existing augmentation strategies is highly task-dependent, we introduce a new method, \textit{TaskAug}, which defines a flexible augmentation policy that is optimized on a per-task basis. We outline an efficient learning algorithm to do so that leverages recent work in nested optimization and implicit differentiation.
In experiments, considering three datasets and eight predictive tasks, we find that TaskAug is competitive with or improves on prior work, and the learned policies shed light on what transformations are most effective for different tasks. We distill key insights from our experimental evaluation, generating a set of best practices for applying data augmentation to ECG prediction problems.
\end{abstract}

\paragraph*{Data and Code Availability}
We use three datasets: two are from Massachusetts General Hospital (MGH) and are not publicly available; the third is PTB-XL~\citep{wagner2020ptb}, which is publicly available on the PhysioNet repository~\citep{Goldberger2000PhysioBankPA}. Code implementing our method is available here: \url{https://github.com/aniruddhraghu/ecg_aug}.

\section{Introduction}
 
Electrocardiography is used widely in medicine as a non-invasive and relatively inexpensive method of measuring the electrical activity in an individual's heart. The output of electrocardiography --- the electrocardiogram (ECG) --- is of great utility to clinicians in diagnosing and monitoring various cardiovascular conditions \citep{salerno2003competency,fesmire1998usefulness, blackburn1960electrocardiogram}. 

In recent years, there has been significant interest in automatically predicting cardiac abnormalities, diseases, and outcomes directly from ECGs using neural network models \citep{hannun2019cardiologist,raghunath2020prediction,gopal20213kg,diamant2021patient,kiyasseh2021clocs,raghu2021learning}.  Although these works demonstrate impressive results, they often require large labelled datasets with paired ECGs and labels to train models. In certain situations, it is challenging to construct such datasets. For example, consider inferring abnormal central hemodynamics (e.g., cardiac output) from the ECG, which is important when monitoring patients with heart failure or pulmonary hypertension \citep{rhcnet}.  Accurate hemodynamics labels are only obtainable through specialized invasive studies \citep{bajorat2006comparison,hiemstra2019diagnostic} and hence it is difficult to obtain large datasets with paired ECGs and hemodynamics variables. 

\begin{figure}[t]
     \centering
         \centering
         \includegraphics[width=0.49\textwidth]{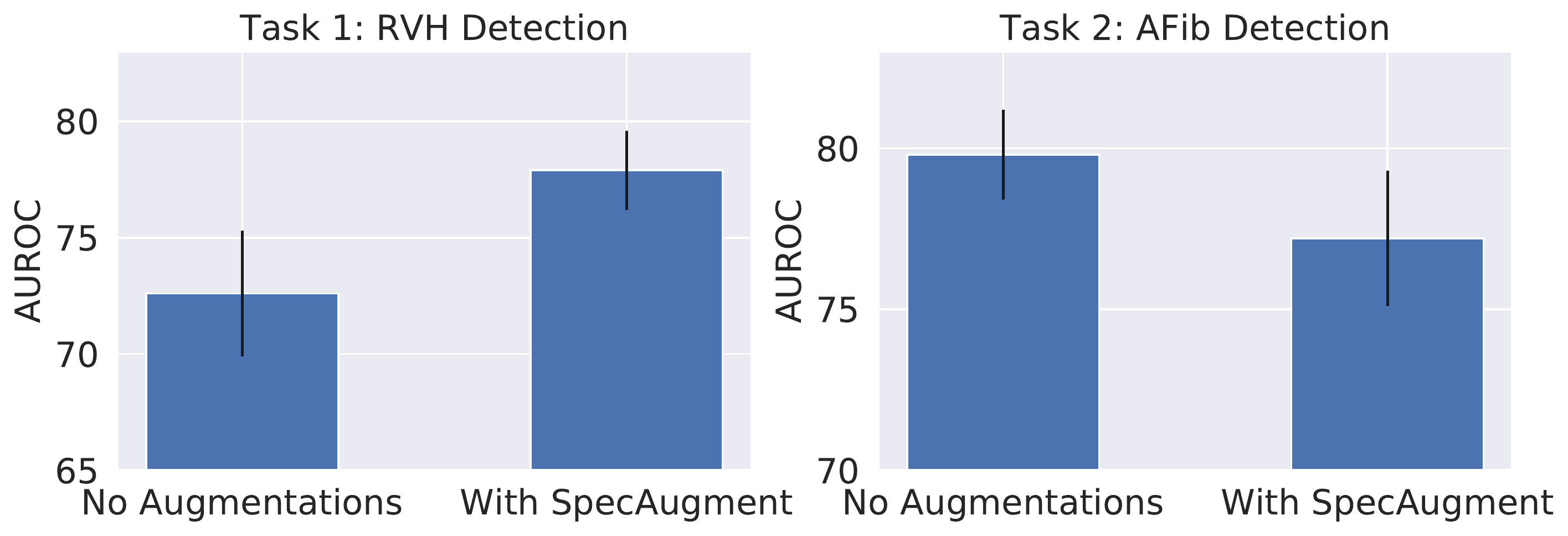}
        \caption{\small \textbf{The effect of data augmentation on ECG prediction tasks is task-dependent.} We examine the mean/standard error of AUROC over 5 runs when applying SpecAugment \citep{park2019specaugment}, a data augmentation method, to two different ECG prediction tasks. We observe performance improvement in one setting (left, Right Ventricular Hypertrophy), and performance reduction in another (right, Atrial Fibrillation).}
        \label{fig:fig1}
\end{figure}

Data augmentation \citep{hataya2020meta,wen2020time,shorten2019survey,iwana2021empirical,cubuk2019autoaugment,cubuk2020randaugment} during training is a useful strategy to improve the predictive performance of models in data-scarce regimes. However, there exists limited work studying data augmentation for ECGs. A key problem with applying standard data augmentations is that fine-grained information within ECGs, such as relative amplitudes of portions of beats, carry predictive signal: augmentations may worsen performance if such predictive signal is destroyed. Furthermore, the effectiveness of data augmentations with ECGs varies on a task-specific basis -- applying the same augmentation for two different tasks could help performance in one case, and hurt performance in another (Figure~\ref{fig:fig1}). 

In this work, we take steps towards addressing these issues. Our contributions are as follows:

\begin{itemize}[leftmargin=*, nosep]
    \item We propose \textit{TaskAug}, a new task-dependent augmentation strategy. TaskAug defines a flexible augmentation policy that is optimized on a per-task basis. We outline an efficient learning algorithm to do so that leverages recent work in nested optimization and implicit differentiation. \citep{lorraine2020optimizing}.
    \item We conduct an empirical study of TaskAug and other augmentation strategies on ECG predictive problems. We consider three datasets and eight different predictive tasks, which cover different classes of cardiac abnormalities. 
    \item We analyze the results from our evaluation, finding that many augmentation strategies do not work well across all tasks. Given its task-specific nature, TaskAug is competitive with or improves on other methods for the problems we examined.
    \item We study the learned TaskAug policies, finding that they offer insights as to what augmentations are most appropriate for different tasks. 
    \item We provide a summary of findings and best practices to assist future studies exploring data augmentation for ECG tasks.
\end{itemize}



\section{Related Work}
\paragraph{Data augmentation for time-series.} Prior research on time-series data augmentation includes: (1) large-scale surveys exploring the impact of augmentation on various downstream modalities \citep{iwana2021empirical,iwana2021time,wen2020time}; and (2) specific methods for particular modalities, including speech signals \citep{park2019specaugment,park2020specaugment}, wearable device signals \citep{um2017data}, and time series forecasting \citep{bandara2021improving,smyl2016data}. 
There is relatively little work exploring how augmentation can impact performance for ECG-based prediction tasks, with prior studies mostly restricted to considering single tasks \citep{hatamian2020effect,banerjee2021afib}. In contrast, in this paper, we evaluate a set of data augmentation methods on many different predictive tasks, studying when and why augmentations may help.
In addition, the data augmentation strategy proposed in this work, \textit{TaskAug}, can be readily adapted to new predictive tasks, unlike in existing works where the methods may be designed for a very specific downstream task.

There also exists related work on using data augmentation for contrastive pre-training with ECGs \citep{gopal20213kg,kiyasseh2021clocs,raghu2021meta,mehari2021Self}. These works are complementary to ours; we focus specifically on supervised learning (rather than contrastive pre-training), and we hypothesize that our proposed augmentation pipeline could be used in these prior methods for improved contrastive learning.

\paragraph{Designing and learning data augmentation policies.} The structure of \textit{TaskAug}, our proposed augmentation strategy, was inspired by related work on flexible data augmentation policies in computer vision \citep{cubuk2019autoaugment,cubuk2020randaugment,hataya2020meta}. We extend these ideas to ECG predictive tasks by (1) selecting appropriate transformations for ECG data, and (2) allowing for class-specific transformation strengths. Since such policies introduce many hyperparameters, we use a bi-level optimization algorithm to enable scalable policy learning \citep{lorraine2020optimizing,raghu2021teaching}.

\section{Problem Setup and Notation}
We focus on supervised binary classification problems from ECG data. 
Let $x \in \mathbb{R}^{12\times T}$ refer to a 12-lead ECG of $T$ samples and $y \in \{0,1\}$ refer to a binary target. We let $\mathcal{D} = \{(x_n, y_n) \}_{n=1}^N$ refer to a dataset of $N$ ECG-label pairs.

Let $f(x;\theta) \rightarrow \nolinebreak \hat{y}$ be a neural network model with parameters $\theta$ that outputs a predicted label $\hat{y}$ given $x$ as input. Network parameters are optimized to minimize the average binary cross entropy loss $\mathcal{L}_{\textit{BCE}}$ on the training dataset $\mathcal{D}^{\textnormal{(train)}}$.

We restrict our study to single label binary classification problems in this work in order to study the effect of data augmentation on a per-task basis. One can extend this to multilabel binary classification by letting $y$ be a vector of several different binary labels and training the network to produce a vector of predictions.

\paragraph{Training with Data Augmentation.} Let $A(x, y;\phi) \rightarrow \nolinebreak \tilde{x}$ refer to a data augmentation function with hyperparameters $\phi$ that takes the input ECG $x$ and its label $y$ and outputs an augmented version $\tilde x$. Note that this formulation implicitly assumes that the augmentation is label preserving, since it does not also change the label $y$. Where relevant, the augmentation hyperparameters $\phi$ may control the strength/probability of applying an augmentation. 

The process of training with data augmentation \footnote{For the SMOTE baseline this process is slightly different; details are in Section \ref{sec:methods}.} amounts to: 
\begin{enumerate}[nosep, leftmargin=*]
    \item Sample a data point and label pair from the training set: $(x,y) \sim \mathcal{D}^{\textnormal{(train)}}$.
    \item Apply the augmentation $A: x \mapsto \tilde{x}$, to transform the original input $x$ to an augmented version $\tilde{x}$.
    \item Use the pair $(\tilde{x}, y)$ in training.
\end{enumerate}

\section{Data Augmentation Methods}
\label{sec:methods}
We now describe the data augmentation methods considered in our experiments. We also present our new, learnable data augmentation method that can be used to find task-specific augmentation policies, and an algorithm to optimize its parameters.

\subsection{Existing Data Augmentation Methods}
We evaluate the following set of existing data augmentation strategies, which includes operations in the signal (time-domain) space, frequency space, and interpolated signal space, providing good coverage of the possible space of augmentations.

\paragraph{Time Masking.} This is a commonly used method in time-series and ECG data augmentation work \citep{iwana2021empirical,gopal20213kg}. We mask out (set to zero) a contiguous fraction $w \in [0,1]$ of the original signal of length $T$, We choose a random starting sample $t_s$ and set all samples $[t_s, t_s +w T] = 0$. 

\paragraph{SpecAugment.} A highly popular method for augmenting speech signals \citep{park2019specaugment,park2020specaugment}. We follow the approach from \citet{kiyasseh2021clocs}, and apply masking (setting components to zero) in the time and frequency domains as follows. We take the Short-Time Fourier Transform (STFT) of the input signal, and independently mask a fraction $w$ of the temporal bins and frequency bins (this involves setting the complex valued entries in these bins to $0 + 0j$). The inverse STFT is then used to map the signal back to the time domain.

\paragraph{Discriminative Guided Warping (DGW).} Introduced in \citet{iwana2021time}, this method uses Dynamic Time Warping (DTW) \citep{muller2007dynamic,berndt1994using} to warp a source ECG to match a representative reference signal that is dissimilar to examples from other classes. 

\paragraph{SMOTE \citep{chawla2002smote}.} A commonly used oversampling strategy, the SMOTE algorithm generates new synthetic examples of the minority class by interpolating minority class samples. Given that many ECG prediction problems are characterized by significant class imbalance, oversampling algorithms are important methods to consider.  
In contrast to the other methods, the SMOTE algorithm generates an augmented dataset prior to any training, based on a predefined training set size, rather than augmenting examples at each training iteration (as presented in Section~\ref{sec:methods}). We set this value to achieve a balanced number of the two classes.


\subsection{\textit{TaskAug}: A New Augmentation Policy}

\paragraph{Motivation.} The approaches mentioned so far are simple to implement and can be effective for various problems; however, they are fairly inflexible, given each individually uses only one or two fixed transformations.
With ECGs, recall that it is unclear on a per-task basis which augmentations may help or worsen performance (Figure \ref{fig:fig1}). Designing a more flexible augmentation strategy that is optimized on a per-task basis could help with this problem, and we now describe such an approach -- \textit{TaskAug}.

\subsubsection{Formalizing TaskAug} 
\label{sec:taskaug-defn}
\paragraph{High-level structure.}  We define a set of operations $\mathcal{S} = \{A_1, \ldots, A_M \}$, each of which is an augmentation function of the form $A_i(x, y; \mu_0, \mu_1)$, where $x$ is the input data point to the augmentation function, $y$ is the label,  and $\{\mu_0, \mu_1\}$ represent the augmentation strengths for datapoints of class label 0 and class label 1 respectively. We separately parameterize the augmentation strengths for each class because transformations may corrupt predictive information in the signal for one class but not the other. 

\begin{figure*}[t]
     \centering
         \centering
         \includegraphics[width=0.65\textwidth]{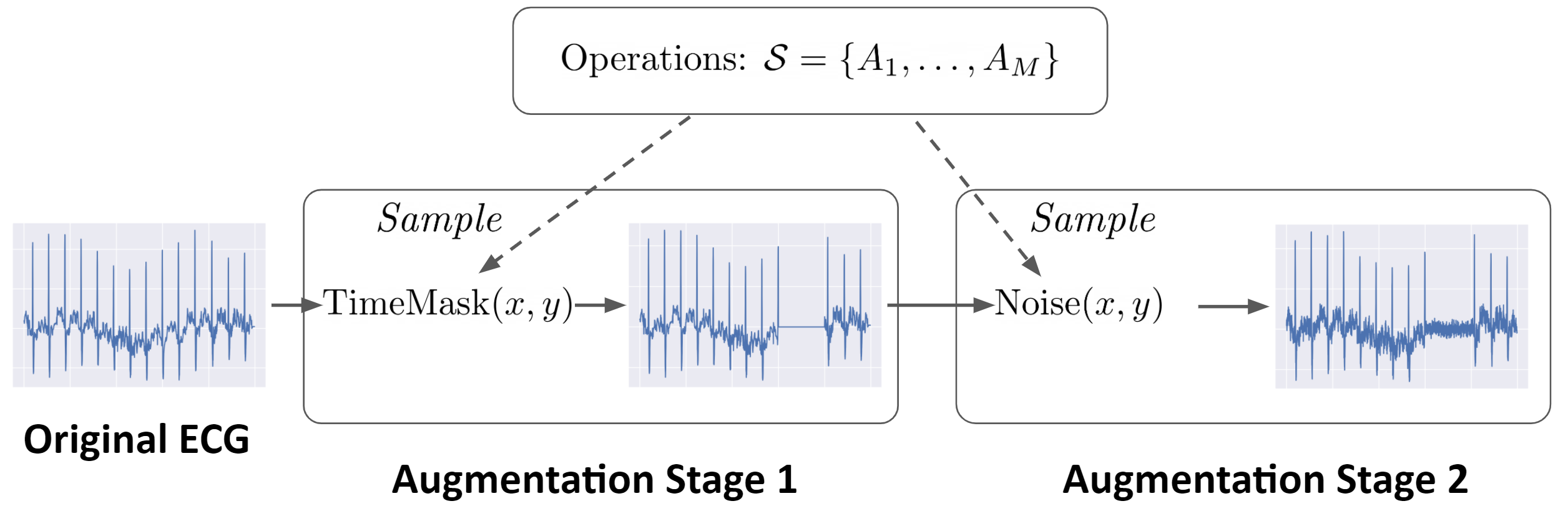}
        \caption{\small \textbf{Structure of TaskAug.} Augmentations to apply are sampled from a set of available operations, and applied in sequence. Here we show an example with $K=2$ stages of augmentation. We omit details relating to the per-class magnitudes and probabilities of sampling for clarity.}
        \label{fig:taskaug}
\end{figure*}

The overall augmentation policy consists of a set of $K$ \textit{stages}, where at each stage we: (1) sample an augmentation function $A_i$ to apply; and (2) apply it to the input signal to that stage. This allows composing combinations of operations in a stochastic manner.  A high-level schematic is shown in Figure~\ref{fig:taskaug}.

\paragraph{Mathematical definition.} The policy is defined following \citet{hataya2020meta}. At each augmentation stage $k \in \{1, \ldots, K\}$ we have a set of operation selection parameters $\mathbf{ \pi}^{(k)} \in [0,1]^{M}$, where $\sum_{i} \pi_i^{(k)} = 1 \ \ \forall k$. Each vector $\pi^{(k)}$ parameterizes a categorical distribution such that each entry $\pi_i^{(k)}$ represents the probability of selecting operation $i$ at augmentation stage $k$. We obtain a reparameterizable sample from this categorical distribution (using the Gumbel-Softmax trick, \citep{jang2016categorical,maddison2016concrete}) at each stage to select the operation to use, as follows:
\begin{align}
    u &\sim \textnormal{Categorical}(\pi^{(k)}) \label{eq:pol1} \quad \texttt{\# Note that $u \in \mathbb{R}^M$}\\
    i &= \arg\max u  \label{eq:pol2} \\
    \tilde{x} &= \frac{u_i}{\texttt{stop\_grad}(u_i)} A_i(x, y; \mu_0, \mu_1) \label{eq:pol3}.
\end{align}
The multiplicative factor $\frac{u_i}{\texttt{stop\_grad}(u_i)}$ allows differentiation w.r.t the operation selection parameters $\pi$.  This enables gradient-based optimization of $\pi$ (see Section \ref{sec:taskaug-opt}). The denominator is necessary because the reparameterized sample from the categorical distribution is not one-hot. Further details are in Appendix \ref{sec:app:augs}.

Suppose a particular augmentation function $A_{i}$ with strength parameters $\mu_0$ and $\mu_1$ is obtained following Eqns \ref{eq:pol1} and \ref{eq:pol2}. Then, denoting the input to this augmentation stage as $x$ with label $y$, the function $A_i$ that computes the augmented output is defined as:  
\begin{align}
    A_i(x, y; \mu_0, \mu_1) = t_i(x; s),
\end{align}
where $t_i$ is the actual transformation applied to the signal (e.g., time masking), and $s$ is the transformation strength, computed as follows: \mbox{$s =  y \mu_1 + (1-y) \mu_0$}. See Appendix \ref{sec:app:augs} for a detailed example of the different steps in applying TaskAug. 

\paragraph{Extension to multiclass and multilabel settings.} Our instantiation of TaskAug is for the binary classification setting, since this is the scenario we consider in our experiments. The formulation can be extended to multiclass/multilabel problems by defining an operation selection probability matrix and strength matrix at each augmentation stage. The operation selection probabilities and operation strengths for a given example are then obtained by taking the matrix product of the relevant parameter matrix and the label vector $y$.

\subsubsection{Optimizing Policy Parameters}
\label{sec:taskaug-opt}

Although the defined policy is flexible, it introduces many new parameters -- for a binary problem, there are $M$ operation selection parameters for the categorical distributions at each stage, and $2$ strength parameters at each stage, resulting in $K \times (2 + M)$ total parameters. Finding effective values for these parameters with random/grid search or Bayesian optimization is computationally expensive since they require training models many times with different parameter settings. We therefore use a gradient-based learning scheme to learn these parameters online. 

We optimize policy parameters to minimize a model's validation loss, which is computed using non-augmented data. Following prior work \citep{lorraine2020optimizing,hataya2020meta,raghu2021teaching}, we alternate gradient updates on the network parameters $\theta$ and the augmentation parameters $\phi$ by iterating the following steps (details and full algorithm in Appendix \ref{sec:app:augs}):
\begin{itemize}[nosep, leftmargin=*]
    \item Optimize the model parameters $\theta$ for $P$ steps: at each step, sample a batch $(x,y)$ of data from $\mathcal{D}^{(\textnormal{train})}$, augment the batch with the augmentation policy to obtain $(\tilde{x}, y)$, compute the predicted label $\hat y$, and update the model parameters using gradient descent: \mbox{$\theta \leftarrow \theta - \eta \nabla \mathcal{L}(y, \hat y)$}.
    \item Compute the validation loss $\mathcal{L}_V$ using an un-augmented batch from the validation dataset.
    \item Perform a gradient update on the augmentation parameters $\phi$. We use the chain rule to re-express the gradient wrt the augmentation parameters:
    
    $$\frac{\partial \mathcal{L}_V}{\partial \phi} = \frac{\partial \mathcal{L}_V}{\partial \theta} \times \frac{\partial \theta}{\partial \phi},$$ 
    
    and compute this as follows. The first term on the RHS is found exactly using straightforward backpropagation; the second term is approximated using the algorithm from \citet{lorraine2020optimizing}, leveraging implicit differentiation for efficient computation (since differentiating through training exactly is too memory-intensive). The augmentation parameters are then updated: \mbox{$\phi \leftarrow \phi - \eta \frac{\partial \mathcal{L}_V}{\partial \phi}$}.
\end{itemize}
By using this algorithm, augmentation parameters are learned on a per-task basis and analyzing the learned parameters may allow us to understand what augmentations are useful for different problems. We return to this in Section~\ref{sec:expts:interpretation}. 

\paragraph{Computational cost.} Optimizing policy parameters in this manner is significantly more computationally efficient than running a grid search over parameter values. With $P=1$, running this algorithm has about $2-3\times$ the computational cost of training without any augmentations.

\begin{table*}[ht!]
\small{
\centering
\centerline{
\begin{tabular}{@{}lllll@{}}
\toprule
\textbf{Dataset}                               & \textbf{Task name}                               & \textbf{Prevalence} &\textbf{Abnormality type}& \textbf{\#ecgs/\#patients} \\ \midrule
\multirow{2}{*}{\begin{tabular}[c]{@{}l@{}}Dataset A \\ \end{tabular}} & Right Ventricular Hypertrophy (RVH)                        & 1\%  &  Structural & 705057/705057              \\
                                               & Atrial Fibrillation (AFib)                                & 5\%                 & Electrical & 705057/705057              \\
 \midrule
\multirow{4}{*}{Dataset B (PTB-XL)}            & Hypertrophy (HYP)                                          & 12\%         & Structural       & 21837/18885                \\
                                               & ST/T Change (STTC)                                       & 22\%         & Ischemia       & 21837/18885                \\
                                               & Conduction Disturbance (CD)                               & 24\%       & Electrical         & 21837/18885                \\
                                               & Myocardial Infarction (MI)                             & 25\%       & Ischemia         & 21837/18885                \\ \midrule
\multirow{2}{*}{\begin{tabular}[c]{@{}l@{}}Dataset C \\ \end{tabular}}& Low Cardiac Ouput (CO)                          & 4\%        & Hemodynamics         & 6290/4051         \\
                                               & \begin{tabular}[c]{@{}l@{}}High Pulmonary Capillary  Wedge \\ Pressure (PCWP)\end{tabular}  & 26\%     & Hemodynamics           & 6290/4051                  \\ \bottomrule
\end{tabular}}}
\caption{\small Summary information about the datasets and tasks considered in our empirical evaluation.}
\label{tab:dataset-task-details}
\end{table*}

\section{Experiments}
We evaluate the data augmentation strategies on ECG prediction tasks. We have two main experimental questions: (1) in what settings can data augmentation be beneficial, and (2) when data augmentation does help, which augmentation strategies are most effective?
To investigate these questions, we consider a range of settings that cover three different 12-lead ECG datasets and eight prediction tasks of varying difficulty, class imbalance, and training set sizes.

\subsection{Experimental Setup}

\subsubsection{Datasets and Tasks} 
We highlight key information about our datasets and tasks here, with a summary in Table~\ref{tab:dataset-task-details}.

\paragraph{Dataset A} is from Massachusetts General Hospital (MGH) and contains paired 12-lead ECGs and labels for different cardiac abnormalities. Of the available labels in the dataset, we select Right Ventricular Hypertrophy (RVH) and Atrial Fibrillation (AFib) as two of the predictive tasks in our evaluation. These were chosen because (1) they have been previously studied as prediction targets from the ECG \citep{couceiro2008detection,rvh-detect}, and (2) they have low positive prevalence: 1\% for RVH, and 5\% for AFib, and therefore help to understand the impact of data augmentation in imbalanced prediction problems.

\paragraph{Dataset B} is PTB-XL~\citep{wagner2020ptb,Goldberger2000PhysioBankPA}, an open-source dataset of 12-lead ECGs. Each ECG has labels for four different categories of cardiac abnormality. This dataset has been used in prior work to evaluate ECG predictive models \citep{gopal20213kg,kiyasseh2021clocs}.

\paragraph{Dataset C} is from the same hospital (MGH) as Dataset A and contains paired ECGs and labels for two hemodynamics parameters, Cardiac Output (CO) and Pulmonary Capillary Wedge Pressure (PCWP). These measures of cardiac health are important in deciding treatment strategies for patients with cardiac disease \citep{yancy20132013,hurst1990heart,solin1999influence}.  Typically, these parameters can only be measured accurately through an invasive cardiac catheterization procedure \citep{bajorat2006comparison,hiemstra2019diagnostic}. As a result, datasets with paired ECGs and hemodynamics measurements are relatively small.  Considering the use of data augmentations to improve model performance in this limited data regime is therefore clinically relevant. We specifically consider inferring abnormally low Cardiac Output, and abnormally high Pulmonary Capillary Wedge Pressure.

Note that the tasks considered cover different classes of cardiac abnormalities: ischemia (MI, STTC), structural (HYP, RVH), electrical (CD, AFib), and abnormal hemodynamics (low CO, high PCWP).

\paragraph{Dataset splitting.} Since the value of data augmentation can depend on the amount of training data, we train on different dataset sizes. For the non-hemodynamic tasks (Datasets A and B), we generate development datasets with 1000, 2500, and 5000 ECGs.
On the more challenging hemodynamics inference tasks (Dataset C), for elevated PCWP, we consider two settings: using a development set of size 1000, and using the full dataset. For low CO, we only use the full dataset, since reducing the dataset size led to poor quality models.

In each setting, we split datasets into development and testing sets on a patient-level (no patient is in both sets).  We split the development set into an 80-20 training-validation split.

\subsubsection{TaskAug Transformations}
Based on prior work in time series and ECG data augmentation \citep{iwana2021empirical,mehari2021Self} we use the following transformations in the TaskAug policy. Mathematical descriptions are in Appendix \ref{sec:app:augs}.
\begin{itemize}[nosep,leftmargin=*]
    \item \textbf{Random temporal warp:}  The signal is warped with a random, diffeomorphic temporal transformation. This is formed by sampling from a zero mean, fixed variance Gaussian at each temporal location in the signal to obtain a velocity field, and then integrating and smoothing (following \citet{balakrishnan2018reg,balakrishnan2019tmi}) to generate a temporal displacement field, which is applied to the signal. The variance is the strength parameter, with higher variance indicating more warping. 
    \item \textbf{Baseline wander:} A low-frequency sinusoidal component is added to the signal, with the amplitude of the sinusoid representing the strength.
    \item \textbf{Gaussian noise:} IID Gaussian noise is added to the signal, with the strength parameter representing the variance of the Gaussian.
    \item \textbf{Magnitude scale:} The signal amplitude is scaled by a number drawn from a scaled uniform distribution, with the scale being the strength parameter.
    \item \textbf{Time mask:} A random contiguous section of the signal is masked out (set to zero).
    \item \textbf{Random temporal displacement:} The entire signal is translated forwards or backwards in time by a random temporal offset, drawn from a uniform distribution scaled by a strength parameter.
\end{itemize}
Note that our instantiation of the augmentation policy could utilize many more operations, but we keep it to this number for simplicity and to assist in interpreting the learned policies.

\subsubsection{Implementation Details}
\paragraph{Network architecture.} We standardize the network architecture to be a 1D convolutional network, based on the ResNet-18 architecture, since prior work has shown architectures of this form to be effective with ECG data \citep{diamant2021patient}. Full architectural details are in the appendix.

\paragraph{Training Details.} On Datasets A and B, all models are trained for 100 epochs, using early stopping based on validation loss. For the hemodynamics inference problems on Dataset C, we train models for 50 epochs with early stopping (since we observed significant overfitting after this point).
We consider 15 random development/testing set splits for Datasets A  and C (lower prevalences for some tasks meant that performance was more variable with fewer runs), and 5 splits for Dataset B.  We train models using the Adam optimizer and a learning rate of 1e-3. This value resulted in stable and effective training across all models (as compared to 1e-4, 5e-4, and 5e-3). As evaluation, we compute the AUROC of the best performing model on the held-out testing set, and report mean/standard error across runs.  We also report results for a baseline (NoAugs) that does not use any data augmentation.

\paragraph{Augmentation Hyperparameters.} In TaskAug, we set the number of augmentation stages to $K=2$ (defined in Section~\ref{sec:taskaug-defn}), following prior work \citep{hataya2020meta}. For the number of model optimization steps $P$ (defined in Section~\ref{sec:taskaug-opt}), we evaluate both $P=1$ and $P=5$, and select the best performing setting based on validation set loss. Further discussion on the choice of $P$ is in Appendix \ref{sec:app:augs}.

For Time Masking and SpecAugment, we search over the masking window, considering $w\in \{0.1, 0.2\}$ for SpecAugment (range based on \citet{kiyasseh2021clocs}) and $w \in \{0.1, 0.2, 0.5\}$ for Time Masking (range based on \citet{gopal20213kg}).

\begin{table*}[h!]
\centering
\small{
\begin{tabular}{@{}lll|llll@{}}
\toprule
          & \multicolumn{2}{c|}{\textbf{Dataset A}}                                                 & \multicolumn{4}{c}{\textbf{Dataset B}}                                                                           \\
\textbf{} & \multicolumn{1}{c}{RVH} & \multicolumn{1}{c|}{AFib}  & \multicolumn{1}{c}{MI} & \multicolumn{1}{c}{HYP} & \multicolumn{1}{c}{STTC} & \multicolumn{1}{c}{CD}  \\ \midrule
NoAugs   & 72.6 $\pm$ 2.7          & {\ul 79.8 $\pm$ 1.4}          & 80.0 $\pm$ 0.8                & \textbf{84.3 $\pm$ 1.4}      & 87.6 $\pm$ 0.8                & 82.2 $\pm$ 0.6          \\
TaskAug   & \textbf{78.4 $\pm$ 1.9} & \textbf{82.8 $\pm$ 1.0}   & \textbf{82.3 $\pm$ 0.5}$^*$       & 83.7 $\pm$ 0.5               & \textbf{87.8 $\pm$ 0.4}       & {\ul 83.1 $\pm$ 0.4}  \\
SMOTE     & 75.9 $\pm$ 1.8          & 79.0 $\pm$ 1.4                       & {\ul 81.2 $\pm$ 0.6}          & 80.4 $\pm$ 0.6               & 87.0 $\pm$ 0.5                & 82.6 $\pm$ 0.8          \\
DGW       & 73.6 $\pm$ 1.7          & 77.4 $\pm$ 1.5                       & 81.1 $\pm$ 0.6                & {\ul 83.9 $\pm$ 0.7}         & 87.5 $\pm$ 0.5                & 81.8 $\pm$ 1.0          \\
SpecAug   & {\ul 77.9 $\pm$ 1.7}    & 77.2 $\pm$ 2.1                       & 81.1 $\pm$ 0.7                & 83.5 $\pm$ 0.8               & {\ul 87.7 $\pm$ 0.4}          & 82.2 $\pm$ 0.7          \\
TimeMask  & 72.8 $\pm$ 2.1          & 77.9 $\pm$ 1.9                       & 81.1 $\pm$ 1.3                & 82.9 $\pm$ 0.7               & 87.7 $\pm$ 0.7                & \textbf{83.8 $\pm$ 1.1}     \\ \bottomrule
\end{tabular}}
\caption{\small \textbf{Augmentation strategies improve AUROC on detecting most cardiac abnormalities in the low-sample regime ($N=1000$), and TaskAug is among the best-performing methods.} Table shows mean and standard error of AUROC (best-performing method bolded,  second best underlined, statistically significant ($p<0.05$) improvement over NoAugs marked $^*$). The impact of augmentations is task-dependent, with some tasks (such as RVH, MI) showing improved performance on average with almost all strategies, and others (HYP) showing no improvement with any strategy. TaskAug is among the best methods across tasks, and improves performance on tasks such as AFib where no other augmentations help.}
\label{tab:arrhythmia-all}
\end{table*}

\subsection{Results}
\label{sec:results}

\begin{table*}[t]
\centering
\small{
\begin{tabular}{@{}lccc@{}}
\toprule
& \multicolumn{3}{c}{\textbf{Dataset C}} \\   
\textbf{}    & {Low CO}      & High PCWP: $N=1000$   & High PCWP: All Data    \\ \midrule
NoAugs      & 65.9 $\pm$ 1.2                              & 66.7 $\pm$ 0.7                              & 74.4 $\pm$ 0.5                              \\
TaskAug      & \underline{68.2 $\pm$ 1.0} & \textbf{67.9 $\pm$ 0.7}    & \textbf{75.1 $\pm$ 0.4}    \\
SMOTE        & 66.0 $\pm$ 1.4                              & 67.2 $\pm$ 0.5                              & 73.6 $\pm$ 0.5                              \\
DGW          & \textbf{68.3 $\pm$ 0.9}    & 66.4 $\pm$ 0.6                              & 74.9 $\pm$ 0.4                              \\
SpecAug  & 66.1 $\pm$ 0.9                              & 66.4 $\pm$ 1.3                              & \underline{75.0 $\pm$ 0.4} \\
TimeMask & 66.8 $\pm$ 1.1                              & \underline{67.3 $\pm$ 0.4} & 74.6 $\pm$ 0.4                              \\ \bottomrule
\end{tabular}}
\caption{\small \textbf{Training with data augmentation improves AUROC on two hemodynamics inference tasks, and TaskAug again is among the best-performing methods.} Table shows mean and standard error of AUROC (best-performing method bolded, second best underlined). All methods are comparable with or improve on the no augmentation baseline for Low CO prediction, possibly because of the low prevalence of the label (4\%). The performance of methods on the High PCWP task is more variable across the two sample sizes. TaskAug obtains improvements in all three settings considered.}
\label{tab:hemo-all}
\end{table*}

\subsubsection{Quantitative results}
\paragraph{Non-hemodynamics tasks.} We first analyze performance of augmentation strategies on the non-hemodynamics tasks. Given that performance improvements are most evident in the lowest sample regimes for both datasets ($N=1000$), we focus on this setting with results shown in Table~\ref{tab:arrhythmia-all}. Results for the higher sample regimes are in the appendix. We summarize key findings here.

\textit{The value of augmentation varies by task}. For some tasks such as RVH and MI, almost all augmentation strategies lead to performance improvements. On other tasks such as STTC and HYP, performance is the same or worse when applying augmentations. The improvement seen with RVH could be due to the fact that it is particularly low prevalence (1\%),  so all augmentation strategies have an oversampling effect and thus boost performance.

\textit{TaskAug performs well on average}. TaskAug almost always improves on the NoAugs baseline, and even boosts performance on some tasks where other augmentations worsen performance (AFib). Although TaskAug does not always result in a statistically significant ($p<0.05$) improvement in AUROC , it is the only method to significantly improve AUPRC over NoAugs on the low-prevalance tasks, RVH and AFib (see Appendix \ref{sec:app:expts}, Table \ref{app:tab:muse-1000-auprc}). \newline
When TaskAug results in lower performance than other augmentation strategies (e.g., TimeMasking for CD), it is still competitive with these methods and never causes a statistically significant reduction in performance compared to other methods.
This suggests that for a new task, it may always be worth using TaskAug to see if performance is boosted. We hypothesise that TaskAug's efficacy is due to its flexible and learned nature, examined in ablation studies (Section \ref{sec:ablation}). 

\textit{Performance improvements are smaller on Dataset B.} The maximum improvement over the NoAugs baseline in Dataset A (5.8\%) is greater than the maximum improvement in Dataset B (2.3\%). We hypothesise two reasons for this. 
Firstly, the prevalence in Dataset B is higher, meaning that augmentations may not have as much of an effect at $N=1000$. We study this in Appendix~\ref{sec:app:expts}, Table~\ref{app:tab:ptbxl-500}, where we examine performance at the $N=500$ data regime for Dataset B, and find that the maximum improvement (obtained with TaskAug for MI) goes up to 4\%. 

Secondly, Dataset A has narrower label definitions than Dataset B, and this affects performance, especially with TaskAug. The HYP, STTC, and CD classes of abnormalities in Dataset B aggregrate many sub-categories together (see Appendix \ref{sec:app:dataset}), and these sub-categories may each benefit from different augmentations. In contrast, the labels in Dataset A are fine-grained, and so TaskAug, which optimizes augmentations on a per-task basis, learns more appropriate augmentation strategies. This hypothesis is supported by the fact that with MI (a more fine-grained label than HYP, CD, and STTC) we observe improvements over the NoAugs baseline (clearly seen in the $N=500$ regime, Appendix~\ref{sec:app:expts}, Table~\ref{app:tab:ptbxl-500}).

\textit{Performance improvements at higher samples are lower}, as seen in the results in Appendix \ref{sec:app:expts}. Augmentations do not worsen performance however, and some tasks (STTC, CD) benefit a small amount, $\sim +1\%$ AUROC.

\paragraph{Hemodynamics tasks.} Table \ref{tab:hemo-all} presents results for performance on the more challenging hemodynamics prediction tasks. All methods are comparable with or improve on the no augmentation baseline for low CO prediction, likely because of the low prevalence of the positive label (4\%). For inferring high PCWP, at both low sample and higher samples, TaskAug obtains improvements in performance (though not significant at the $p<0.05$ level); however, other methods do not consistently improve on the no augmentation baseline. Although improvements in AUROC are not statistically significant, we observe significant improvements with TaskAug in AUPRC for low CO detection (see Appendix \ref{sec:app:expts}, Table \ref{app:tab:hemo-auprc}).
Again, we see that the benefit of augmentation varies with the task, prevalance, and dataset size, and that TaskAug is better than or competitive with other strategies.

\begin{figure*}[t!]
\floatconts
{fig:afib-policy}
{\caption{\small \textbf{The TaskAug policy for Atrial Fibrillation detection.} We focus on the probability of selecting each transformation in both augmentation stages (left) and the optimized temporal warp strengths in the first stage (right). We show the mean/standard error of these optimized policy parameters over 15 runs. Given the characteristic features of AFib (e.g., irregular R-R interval), Time Masking is likely to be label preserving and therefore it is sensible that it has a high probability of selection. The temporal warp strength for positive samples is higher than that for negative samples, which makes sense since time warping a negative sample too strongly could change its label.}}
{%
     \subfigure[\small Operation selection probabilities]{\label{fig:afib-policy-probs}%
        \includegraphics[width=0.65\linewidth]{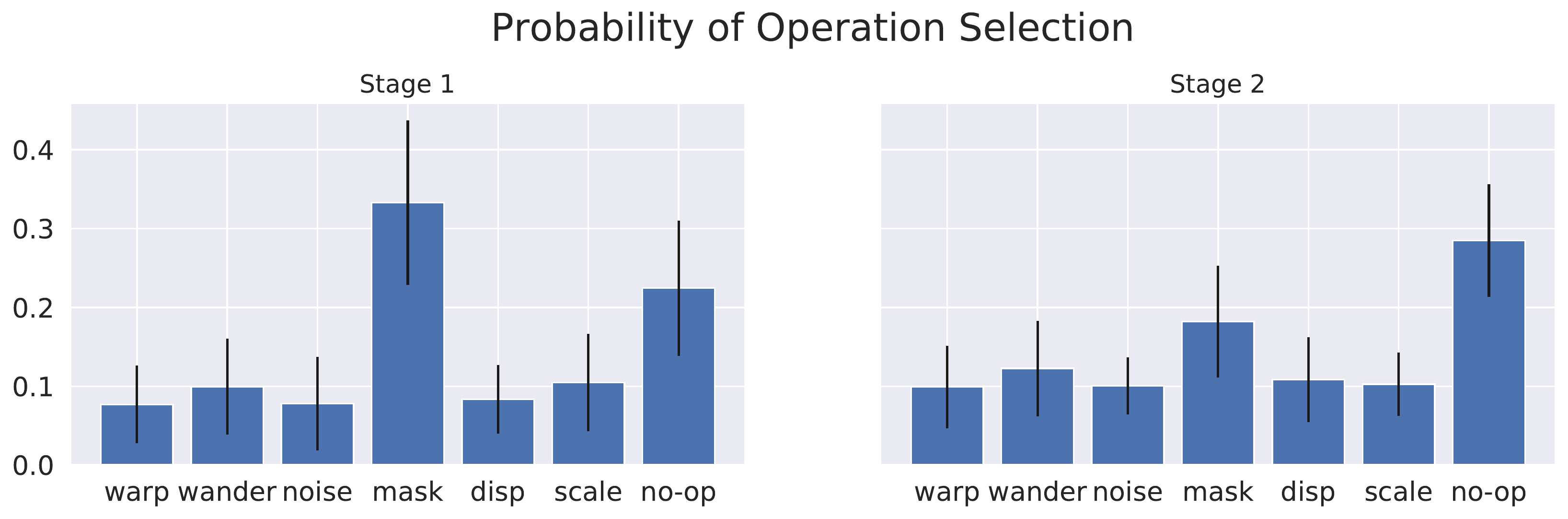}}%
        \qquad
     \subfigure[\small Warp strengths]{\label{fig:afib-policy-warps}%
        \includegraphics[width=0.29\linewidth]{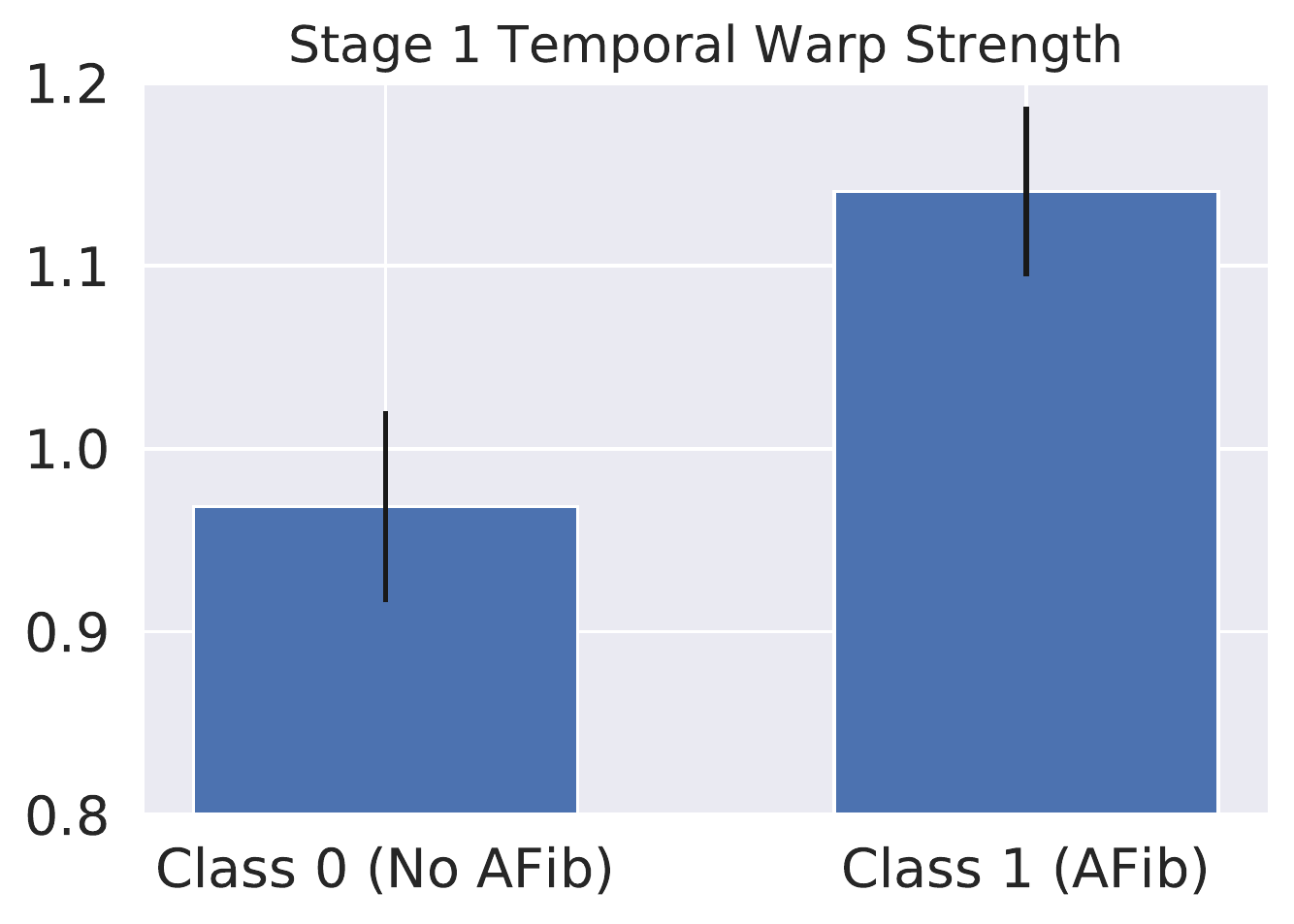}}
}
\end{figure*}

\begin{figure*}[t!]
\floatconts
{fig:pcwp-policy}
{\caption{\small \textbf{The TaskAug policy for detecting elevated Pulmonary Capillary Wedge Pressure.} We focus on the probability of selecting each transformation in both augmentation stages (left) and the optimized magnitude scaling strengths in the second stage (right).  We show the mean/standard error of these optimized policy parameters over 15 runs.  There exists little domain knowledge about what features in the ECG may encode elevated PCWP, so examining the learned augmentations here could provide hypotheses of invariances in the data. Of interest is that the positive class is augmented with stronger magnitude scaling  than the negative class, suggesting that scaling negative examples could affect their labels.}}
{%
     \subfigure[\small Operation selection probabilities]{\label{fig:pcwp-policy-probs}%
        \includegraphics[width=0.65\linewidth]{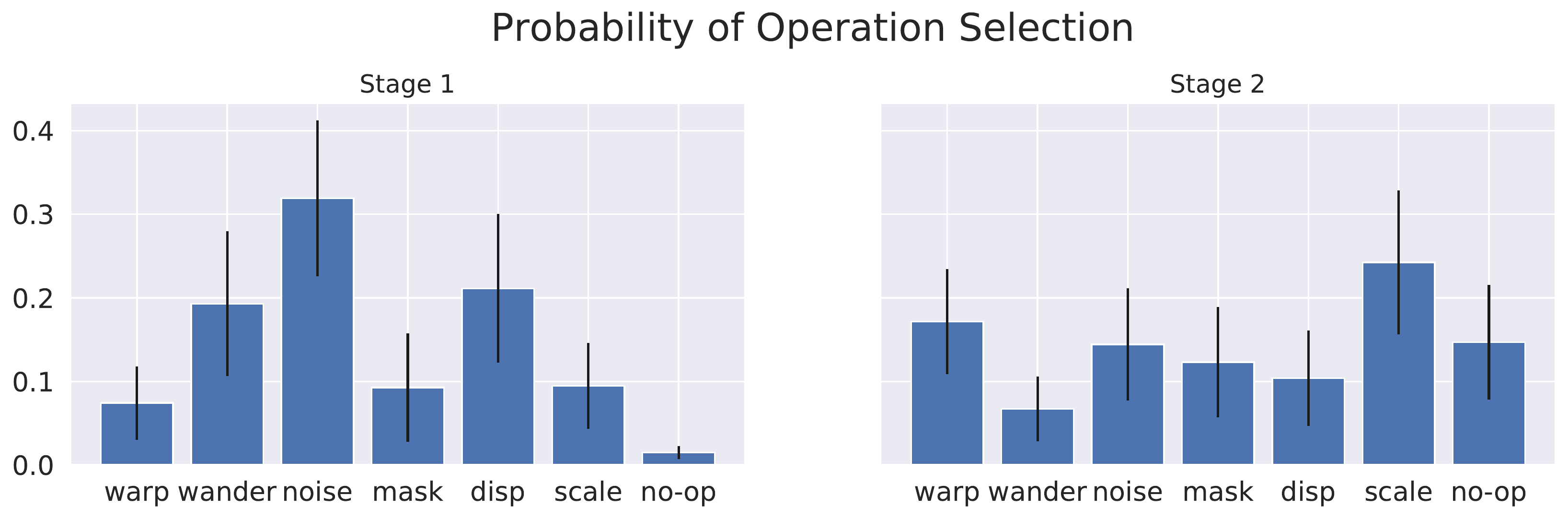}}%
        \qquad
     \subfigure[\small Magnitude scale strength]{\label{fig:pcwp-policy-mags}%
        \includegraphics[width=0.29\linewidth]{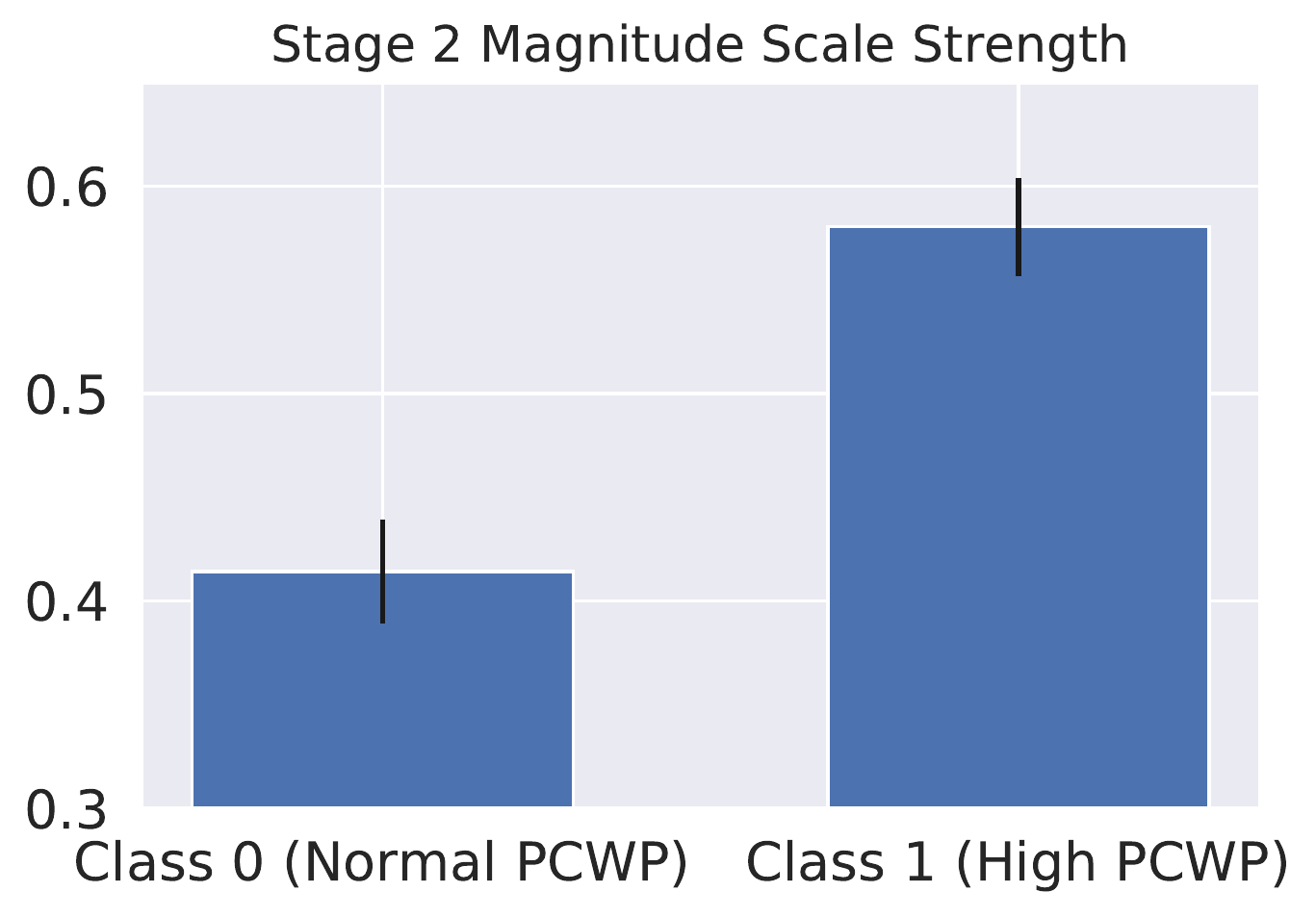}}
}
\end{figure*}

\subsubsection{Analyzing learned policies}
\label{sec:expts:interpretation}
We analyze the learned policies for three of the predictive tasks: AFib, PCWP, and RVH (appendix). 

\paragraph{AFib, Figure~\ref{fig:afib-policy}.} We see that time mask has a high probability of selection (Figure~\ref{fig:afib-policy-probs}). Since AFib is characterized in the ECG by an irregular R peak-R peak interval \citep{couceiro2008detection}, which is often present regardless of which section of ECG is selected, time masking is likely label preserving, and is a sensible choice. Considering the learned time warp strength in Figure~\ref{fig:afib-policy-warps}, we observe that signals labelled negative for AFib are warped less strongly than those with AFib, again sensible since time warping may affect the label of a signal and introduce AFib in a signal where it was not originally present.

\begin{figure}[t]
     \centering
         \centering
         \includegraphics[width=\linewidth]{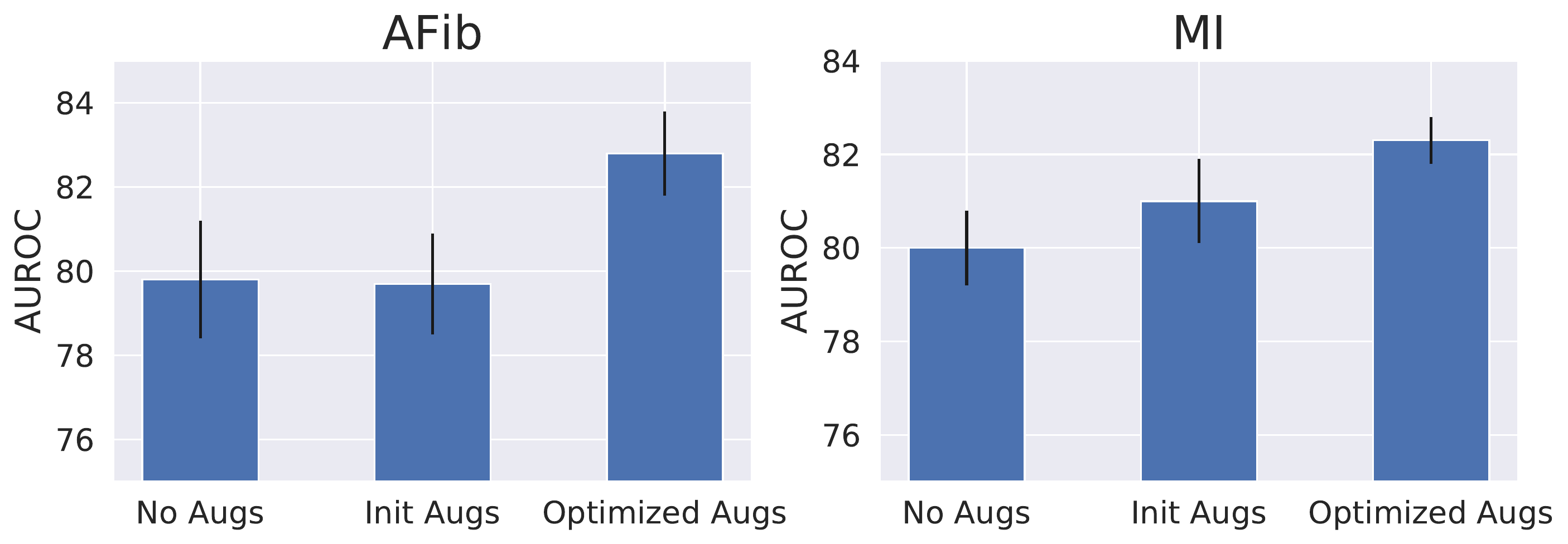}
        \caption{\small \textbf{Optimizing the TaskAug policy parameters results in performance improvements.} We show the mean/standard error of AUROC over 15 runs for AFib and over 5 runs for MI.  Without optimizing policy parameters (InitAug), performance is comparable to not using augmentations at all, indicating the importance of learning the policy parameters.}
        \label{fig:ablation}
\end{figure}

\paragraph{PCWP, Figure~\ref{fig:pcwp-policy}.} We have limited domain understanding of what augmentations may be label preserving and help model performance, since detecting high PCWP from ECGs is not something clinicians are typically able to do \citep{rhcnet}. Analyzing the augmentations could provide hypotheses about what features in the data encode the class label. Noise, displacement, and baseline wander all obtain higher weight in the first stage, and scaling obtains higher weight in the second stage. The high weight assigned to noise could be to help the model build invariance to it, and not use it as a predictive aspect of the signal. Studying the magnitude scaling in Figure~\ref{fig:pcwp-policy-mags}, we see positive examples are scaled significantly more than negative examples. It is possible that negative examples are more sensitive to scale, and scaling them pushes them into positive example space. The positive examples may have more variance in scaling, and thus scaling them further has less of an effect.

\subsubsection{Ablation Studies}
\label{sec:ablation}
\paragraph{How much does optimizing augmentations help?} Our results show that TaskAug offers improvements in performance. In Figure~\ref{fig:ablation}, we examine how the actual optimization of the augmentation policy parameters (operation selection probabilities and magnitudes, Section~\ref{sec:taskaug-defn}) affects performance, considering the AFib and MI detection tasks and $N=1000$. We compare the performance of optimizing the policy parameters vs. keeping them fixed at their initialized values and training. We observe improvements in performance through the optimization process, suggesting that it is not only the range of augmentations that leads to improved performance, but also the optimization of the policy parameters. In Appendix \ref{sec:app:expts}, we study this at different dataset sizes and find that performance is improved by optimization at each size.

\begin{figure}[t]
     \centering
         \centering
         \includegraphics[width=\linewidth]{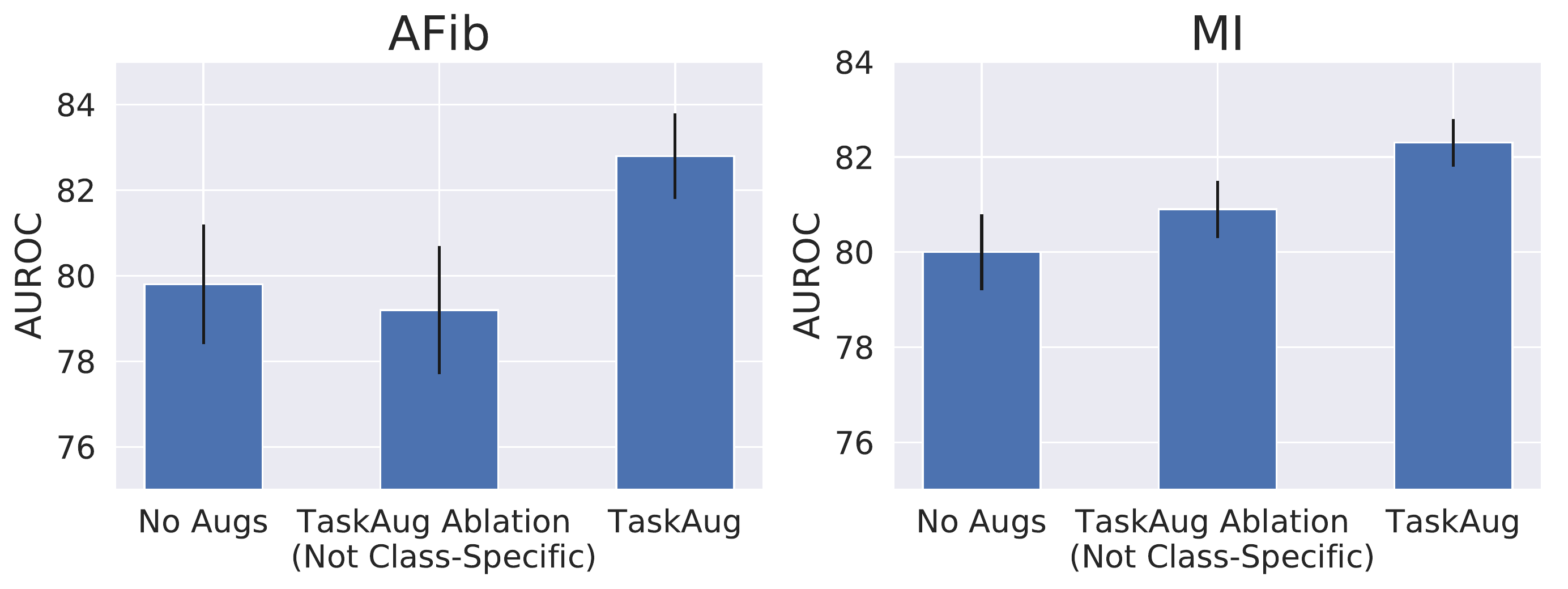}
        \caption{\small \textbf{Class-specific magnitude parameters in TaskAug lead to improvements in performance.} We show the mean/standard error of AUROC over 15 runs for AFib and over 5 runs for MI. This is particularly true for tasks such as AFib where some operations may not be label preserving.}
        \label{fig:ablation2}
\end{figure}

\paragraph{How much do class-specific magnitudes help?} TaskAug instantiates magnitude parameters for the augmentation operations on a per-class basis, as described in Section~\ref{sec:taskaug-defn}, allowing positive and negative examples to be augmented differently. We examine this further, considering the AFib and MI detection tasks and $N=1000$. We compare performance using class-specific magnitude parameters (the positive and negative examples have independent augmentation magnitudes $\mu_1$ and $\mu_0$) vs. using global magnitude parameters (the positive and negative examples are forced to have the same augmentation magnitude: $\mu = \mu_0 = \mu_1$). 
Results are shown in Figure~\ref{fig:ablation2}. We observe noticeable improvements in performance with class-specific magnitude parameters, demonstrating the importance of independently specifying magnitudes for the two classes. In Appendix \ref{sec:app:expts}, we study this at different dataset sizes and find that performance is improved at each size.

\subsubsection{Summary and best practices}
\begin{itemize}[nosep, leftmargin=*]
    \item Training with data augmentations does not always improve model performance, and may even hurt it. The impact of augmentation depends on nature of the task, positive class prevalence, and dataset size. 
    \item Augmentations are most often useful in the low-sample regime. Where the prevalence is particularly low  (see results for RVH detection) various augmentation strategies improve performance, perhaps by functioning as a form of oversampling.
    \item Data augmentations do not always improve performance at high sample sizes, but do not hurt it.
    \item TaskAug, our proposed augmentation strategy, is the most effective method on average, and could therefore be the first augmentation strategy one tries on a new ECG prediction problem. TaskAug defines a flexible augmentation policy that is optimized on a task-dependent basis, which directly contributes to its effectiveness.
    \item TaskAug also offers insights as to what augmentations are most effective for a given problem, which could be useful in novel prediction tasks (e.g., hemodynamics inference) to suggest what aspects of the ECG determine the class label.
\end{itemize}

\section{Conclusion}
In this work, we studied the use of data augmentation for prediction problems from 12-lead electrocardiograms (ECGs). We outlined \textit{TaskAug}, a new, learnable data-augmentation strategy for ECGs, and conducted an empirical study of this method and several existing augmentation strategies. 

In our experimental evaluation on three ECG datasets and eight distinct predictive tasks, we find that data augmentation is not always helpful for ECG prediction problems, and for some tasks may worsen performance. Augmentations can be most helpful in the low-sample regime, and specifically when the prevalence of the positive class is low. Our proposed learnable augmentation strategy, TaskAug, was among the strongest performing methods in all tasks. TaskAug augmentation policies are additionally interpretable, providing insight as to what transformations are most important for different problems. Future work could consider applying TaskAug to other settings (e.g., multiview contrastive learning) and modalities (e.g., EEGs) where flexible augmentation policies may be useful and could be interpreted to provide scientific insight.
\clearpage

\section*{Institutional Review Board (IRB)}
This study was approved by the Institutional Review Board (IRB) at Massachusetts General Hospital (protocol 2020P000132).

\bibliography{references}

\begin{thebibliography}{45}
\providecommand{\natexlab}[1]{#1}
\providecommand{\url}[1]{\texttt{#1}}
\expandafter\ifx\csname urlstyle\endcsname\relax
  \providecommand{\doi}[1]{doi: #1}\else
  \providecommand{\doi}{doi: \begingroup \urlstyle{rm}\Url}\fi

\bibitem[Bajorat et~al.(2006)Bajorat, Hofmockel, Vagts, Janda, Pohl, Beck, and
  Noeldge-Schomburg]{bajorat2006comparison}
J~Bajorat, R~Hofmockel, DA~Vagts, M~Janda, B~Pohl, C~Beck, and
  G~Noeldge-Schomburg.
\newblock Comparison of invasive and less-invasive techniques of cardiac output
  measurement under different haemodynamic conditions in a pig model.
\newblock \emph{European journal of anaesthesiology}, 23\penalty0 (1):\penalty0
  23--30, 2006.

\bibitem[Balakrishnan et~al.(2018)Balakrishnan, Zhao, Sabuncu, Guttag, and
  Dalca]{balakrishnan2018reg}
Guha Balakrishnan, Amy Zhao, Mert Sabuncu, John Guttag, and Adrian~V. Dalca.
\newblock An unsupervised learning model for deformable medical image
  registration.
\newblock \emph{CVPR: Computer Vision and Pattern Recognition}, pages
  9252--9260, 2018.

\bibitem[Balakrishnan et~al.(2019)Balakrishnan, Zhao, Sabuncu, Guttag, and
  Dalca]{balakrishnan2019tmi}
Guha Balakrishnan, Amy Zhao, Mert Sabuncu, John Guttag, and Adrian~V. Dalca.
\newblock Voxelmorph: A learning framework for deformable medical image
  registration.
\newblock \emph{IEEE TMI: Transactions on Medical Imaging}, 38:\penalty0
  1788--1800, 2019.

\bibitem[Bandara et~al.(2021)Bandara, Hewamalage, Liu, Kang, and
  Bergmeir]{bandara2021improving}
Kasun Bandara, Hansika Hewamalage, Yuan-Hao Liu, Yanfei Kang, and Christoph
  Bergmeir.
\newblock Improving the accuracy of global forecasting models using time series
  data augmentation.
\newblock \emph{Pattern Recognition}, 120:\penalty0 108148, 2021.

\bibitem[Banerjee and Ghose(2021)]{banerjee2021afib}
Rohan Banerjee and Avik Ghose.
\newblock Synthesis of realistic ecg waveforms using a composite generative
  adversarial network for classification of atrial fibrillation.
\newblock In \emph{2021 29th European Signal Processing Conference (EUSIPCO)},
  pages 1145--1149, 2021.
\newblock \doi{10.23919/EUSIPCO54536.2021.9616079}.

\bibitem[Berndt and Clifford(1994)]{berndt1994using}
Donald~J Berndt and James Clifford.
\newblock Using dynamic time warping to find patterns in time series.
\newblock In \emph{KDD workshop}, volume~10, pages 359--370. Seattle, WA, USA:,
  1994.

\bibitem[Blackburn et~al.(1960)Blackburn, Keys, Simonson, Rautaharju, and
  Punsar]{blackburn1960electrocardiogram}
Henry Blackburn, Ancel Keys, Ernst Simonson, Pentti Rautaharju, and Sven
  Punsar.
\newblock The electrocardiogram in population studies: a classification system.
\newblock \emph{Circulation}, 21\penalty0 (6):\penalty0 1160--1175, 1960.

\bibitem[Chawla et~al.(2002)Chawla, Bowyer, Hall, and
  Kegelmeyer]{chawla2002smote}
Nitesh~V Chawla, Kevin~W Bowyer, Lawrence~O Hall, and W~Philip Kegelmeyer.
\newblock Smote: synthetic minority over-sampling technique.
\newblock \emph{Journal of artificial intelligence research}, 16:\penalty0
  321--357, 2002.

\bibitem[Couceiro et~al.(2008)Couceiro, Carvalho, Henriques, Antunes, Harris,
  and Habetha]{couceiro2008detection}
Ricardo Couceiro, Paulo Carvalho, Jorge Henriques, Manuel Antunes, Matthew
  Harris, and J{\"o}rg Habetha.
\newblock Detection of atrial fibrillation using model-based ecg analysis.
\newblock In \emph{2008 19th International Conference on Pattern Recognition},
  pages 1--5. IEEE, 2008.

\bibitem[Cubuk et~al.(2019)Cubuk, Zoph, Mane, Vasudevan, and
  Le]{cubuk2019autoaugment}
Ekin~D Cubuk, Barret Zoph, Dandelion Mane, Vijay Vasudevan, and Quoc~V Le.
\newblock Autoaugment: Learning augmentation strategies from data.
\newblock In \emph{Proceedings of the IEEE/CVF Conference on Computer Vision
  and Pattern Recognition}, pages 113--123, 2019.

\bibitem[Cubuk et~al.(2020)Cubuk, Zoph, Shlens, and Le]{cubuk2020randaugment}
Ekin~D Cubuk, Barret Zoph, Jonathon Shlens, and Quoc~V Le.
\newblock Randaugment: Practical automated data augmentation with a reduced
  search space.
\newblock In \emph{Proceedings of the IEEE/CVF Conference on Computer Vision
  and Pattern Recognition Workshops}, pages 702--703, 2020.

\bibitem[Diamant et~al.(2021)Diamant, Reinertsen, Song, Aguirre, Stultz, and
  Batra]{diamant2021patient}
Nathaniel Diamant, Erik Reinertsen, Steven Song, Aaron Aguirre, Collin Stultz,
  and Puneet Batra.
\newblock Patient contrastive learning: a performant, expressive, and practical
  approach to ecg modeling.
\newblock 2021.

\bibitem[Fesmire et~al.(1998)Fesmire, Percy, Bardoner, Wharton, and
  Calhoun]{fesmire1998usefulness}
Francis~M Fesmire, Robert~F Percy, Jim~B Bardoner, David~R Wharton, and Frank~B
  Calhoun.
\newblock Usefulness of automated serial 12-lead ecg monitoring during the
  initial emergency department evaluation of patients with chest pain.
\newblock \emph{Annals of emergency medicine}, 31\penalty0 (1):\penalty0 3--11,
  1998.

\bibitem[Goldberger et~al.(2000)Goldberger, Amaral, Glass, Hausdorff, Ivanov,
  Mark, Mietus, Moody, Peng, and Stanley]{Goldberger2000PhysioBankPA}
A.~Goldberger, L.~A. Amaral, L.~Glass, Jeffrey~M. Hausdorff, P.~Ivanov,
  R.~Mark, J.~Mietus, G.~Moody, C.~Peng, and H.~Stanley.
\newblock {PhysioBank, PhysioToolkit, and PhysioNet}: components of a new
  research resource for complex physiologic signals.
\newblock \emph{Circulation}, 101 23:\penalty0 E215--20, 2000.

\bibitem[Gopal et~al.(2021)Gopal, Han, Raghupathi, Ng, Tison, and
  Rajpurkar]{gopal20213kg}
Bryan Gopal, Ryan~W. Han, Gautham Raghupathi, Andrew~Y. Ng, Geoffrey~H. Tison,
  and Pranav Rajpurkar.
\newblock 3kg: Contrastive learning of 12-lead electrocardiograms using
  physiologically-inspired augmentations.
\newblock 2021.

\bibitem[Hannun et~al.(2019)Hannun, Rajpurkar, Haghpanahi, Tison, Bourn,
  Turakhia, and Ng]{hannun2019cardiologist}
Awni~Y Hannun, Pranav Rajpurkar, Masoumeh Haghpanahi, Geoffrey~H Tison, Codie
  Bourn, Mintu~P Turakhia, and Andrew~Y Ng.
\newblock Cardiologist-level arrhythmia detection and classification in
  ambulatory electrocardiograms using a deep neural network.
\newblock \emph{Nature medicine}, 25\penalty0 (1):\penalty0 65--69, 2019.

\bibitem[Hatamian et~al.(2020)Hatamian, Ravikumar, Vesal, Kemeth, Struck, and
  Maier]{hatamian2020effect}
Faezeh~Nejati Hatamian, Nishant Ravikumar, Sulaiman Vesal, Felix~P Kemeth,
  Matthias Struck, and Andreas Maier.
\newblock The effect of data augmentation on classification of atrial
  fibrillation in short single-lead ecg signals using deep neural networks.
\newblock In \emph{ICASSP 2020-2020 IEEE International Conference on Acoustics,
  Speech and Signal Processing (ICASSP)}, pages 1264--1268. IEEE, 2020.

\bibitem[Hataya et~al.(2020)Hataya, Zdenek, Yoshizoe, and
  Nakayama]{hataya2020meta}
Ryuichiro Hataya, Jan Zdenek, Kazuki Yoshizoe, and Hideki Nakayama.
\newblock Meta approach to data augmentation optimization.
\newblock \emph{arXiv preprint arXiv:2006.07965}, 2020.

\bibitem[He et~al.(2016)He, Zhang, Ren, and Sun]{he2016deep}
Kaiming He, Xiangyu Zhang, Shaoqing Ren, and Jian Sun.
\newblock Deep residual learning for image recognition.
\newblock In \emph{Conference on Computer Vision and Pattern Recognition},
  pages 770--778, 2016.

\bibitem[Hiemstra et~al.(2019)Hiemstra, Koster, Wiersema, Hummel, van~der
  Harst, Snieder, Eck, Kaufmann, Scheeren, Perner,
  et~al.]{hiemstra2019diagnostic}
Bart Hiemstra, Geert Koster, Renske Wiersema, Yoran~M Hummel, Pim van~der
  Harst, Harold Snieder, Ruben~J Eck, Thomas Kaufmann, Thomas~WL Scheeren,
  Anders Perner, et~al.
\newblock The diagnostic accuracy of clinical examination for estimating
  cardiac index in critically ill patients: the simple intensive care
  studies-i.
\newblock \emph{Intensive care medicine}, 45\penalty0 (2):\penalty0 190--200,
  2019.

\bibitem[Hurst et~al.(1990)Hurst, Rackley, Sonnenblick, and
  Wenger]{hurst1990heart}
J~Hurst, C~Rackley, E~Sonnenblick, and N~Wenger.
\newblock \emph{The heart, arteries and veins}, volume~1.
\newblock McGraw-Hill, 1990.

\bibitem[Iwana and Uchida(2021{\natexlab{a}})]{iwana2021empirical}
Brian~Kenji Iwana and Seiichi Uchida.
\newblock An empirical survey of data augmentation for time series
  classification with neural networks.
\newblock \emph{Plos one}, 16\penalty0 (7):\penalty0 e0254841,
  2021{\natexlab{a}}.

\bibitem[Iwana and Uchida(2021{\natexlab{b}})]{iwana2021time}
Brian~Kenji Iwana and Seiichi Uchida.
\newblock Time series data augmentation for neural networks by time warping
  with a discriminative teacher.
\newblock In \emph{2020 25th International Conference on Pattern Recognition
  (ICPR)}, pages 3558--3565. IEEE, 2021{\natexlab{b}}.

\bibitem[Jang et~al.(2016)Jang, Gu, and Poole]{jang2016categorical}
Eric Jang, Shixiang Gu, and Ben Poole.
\newblock Categorical reparameterization with gumbel-softmax.
\newblock \emph{arXiv preprint arXiv:1611.01144}, 2016.

\bibitem[Kiyasseh et~al.(2021)Kiyasseh, Zhu, and Clifton]{kiyasseh2021clocs}
Dani Kiyasseh, Tingting Zhu, and David~A Clifton.
\newblock Clocs: Contrastive learning of cardiac signals across space, time,
  and patients.
\newblock In \emph{International Conference on Machine Learning}, pages
  5606--5615. PMLR, 2021.

\bibitem[Lin and Lu(2020)]{rvh-detect}
Gen-Min Lin and Henry Horng-Shing Lu.
\newblock A 12-lead ecg-based system with physiological parameters and machine
  learning to identify right ventricular hypertrophy in young adults.
\newblock \emph{IEEE Journal of Translational Engineering in Health and
  Medicine}, 8:\penalty0 1--10, 2020.
\newblock \doi{10.1109/JTEHM.2020.2996370}.

\bibitem[Lorraine et~al.(2020)Lorraine, Vicol, and
  Duvenaud]{lorraine2020optimizing}
Jonathan Lorraine, Paul Vicol, and David Duvenaud.
\newblock Optimizing millions of hyperparameters by implicit differentiation.
\newblock In \emph{International Conference on Artificial Intelligence and
  Statistics}, pages 1540--1552. PMLR, 2020.

\bibitem[Maddison et~al.(2016)Maddison, Mnih, and Teh]{maddison2016concrete}
Chris~J Maddison, Andriy Mnih, and Yee~Whye Teh.
\newblock The concrete distribution: A continuous relaxation of discrete random
  variables.
\newblock \emph{arXiv preprint arXiv:1611.00712}, 2016.

\bibitem[Mehari and Strodthoff(2021)]{mehari2021Self}
Temesgen Mehari and Nils Strodthoff.
\newblock Self-supervised representation learning from 12-lead ecg data.
\newblock \emph{arXiv preprint 2103.12676}, 2021.

\bibitem[M{\"u}ller(2007)]{muller2007dynamic}
Meinard M{\"u}ller.
\newblock Dynamic time warping.
\newblock \emph{Information retrieval for music and motion}, pages 69--84,
  2007.

\bibitem[Park et~al.(2019)Park, Chan, Zhang, Chiu, Zoph, Cubuk, and
  Le]{park2019specaugment}
Daniel~S Park, William Chan, Yu~Zhang, Chung-Cheng Chiu, Barret Zoph, Ekin~D
  Cubuk, and Quoc~V Le.
\newblock Specaugment: A simple data augmentation method for automatic speech
  recognition.
\newblock \emph{arXiv preprint arXiv:1904.08779}, 2019.

\bibitem[Park et~al.(2020)Park, Zhang, Chiu, Chen, Li, Chan, Le, and
  Wu]{park2020specaugment}
Daniel~S Park, Yu~Zhang, Chung-Cheng Chiu, Youzheng Chen, Bo~Li, William Chan,
  Quoc~V Le, and Yonghui Wu.
\newblock Specaugment on large scale datasets.
\newblock In \emph{ICASSP 2020-2020 IEEE International Conference on Acoustics,
  Speech and Signal Processing (ICASSP)}, pages 6879--6883. IEEE, 2020.

\bibitem[Raghu et~al.(2021{\natexlab{a}})Raghu, Guttag, Young, Pomerantsev,
  Dalca, and Stultz]{raghu2021learning}
Aniruddh Raghu, John Guttag, Katherine Young, Eugene Pomerantsev, Adrian~V
  Dalca, and Collin~M Stultz.
\newblock Learning to predict with supporting evidence: applications to
  clinical risk prediction.
\newblock In \emph{Proceedings of the Conference on Health, Inference, and
  Learning}, pages 95--104, 2021{\natexlab{a}}.

\bibitem[Raghu et~al.(2021{\natexlab{b}})Raghu, Lorraine, Kornblith, McDermott,
  and Duvenaud]{raghu2021meta}
Aniruddh Raghu, Jonathan Lorraine, Simon Kornblith, Matthew McDermott, and
  David~K Duvenaud.
\newblock Meta-learning to improve pre-training.
\newblock \emph{Advances in Neural Information Processing Systems}, 34,
  2021{\natexlab{b}}.

\bibitem[Raghu et~al.(2021{\natexlab{c}})Raghu, Raghu, Kornblith, Duvenaud, and
  Hinton]{raghu2021teaching}
Aniruddh Raghu, Maithra Raghu, Simon Kornblith, David Duvenaud, and Geoffrey
  Hinton.
\newblock Teaching with commentaries.
\newblock In \emph{International Conference on Learning Representations},
  2021{\natexlab{c}}.

\bibitem[Raghunath et~al.(2020)Raghunath, Cerna, Jing, Stough, Hartzel, Leader,
  Kirchner, Stumpe, Hafez, Nemani, et~al.]{raghunath2020prediction}
Sushravya Raghunath, Alvaro E~Ulloa Cerna, Linyuan Jing, Joshua Stough,
  Dustin~N Hartzel, Joseph~B Leader, H~Lester Kirchner, Martin~C Stumpe, Ashraf
  Hafez, Arun Nemani, et~al.
\newblock Prediction of mortality from 12-lead electrocardiogram voltage data
  using a deep neural network.
\newblock \emph{Nature medicine}, 26\penalty0 (6):\penalty0 886--891, 2020.

\bibitem[Salerno et~al.(2003)Salerno, Alguire, and
  Waxman]{salerno2003competency}
Stephen~M Salerno, Patrick~C Alguire, and Herbert~S Waxman.
\newblock Competency in interpretation of 12-lead electrocardiograms: a summary
  and appraisal of published evidence.
\newblock \emph{Annals of Internal Medicine}, 138\penalty0 (9):\penalty0
  751--760, 2003.

\bibitem[Schlesinger et~al.(2021)Schlesinger, Diamant, Raghu, Reinertsen,
  Young, Batra, Pomerantsev, and Stultz]{rhcnet}
Daphne Schlesinger, Nathaniel Diamant, Aniruddh Raghu, Erik Reinertsen,
  Katherine Young, Puneet Batra, Eugene Pomerantsev, and Collin~M. Stultz.
\newblock A deep learning model for inferring elevated pulmonary capillary
  wedge pressures from the 12-lead electrocardiogram.
\newblock 2021.

\bibitem[Shorten and Khoshgoftaar(2019)]{shorten2019survey}
Connor Shorten and Taghi~M Khoshgoftaar.
\newblock A survey on image data augmentation for deep learning.
\newblock \emph{Journal of Big Data}, 6\penalty0 (1):\penalty0 1--48, 2019.

\bibitem[Smyl and Kuber(2016)]{smyl2016data}
Slawek Smyl and Karthik Kuber.
\newblock Data preprocessing and augmentation for multiple short time series
  forecasting with recurrent neural networks.
\newblock In \emph{36th International Symposium on Forecasting}, 2016.

\bibitem[Solin et~al.(1999)Solin, Bergin, Richardson, Kaye, Walters, and
  Naughton]{solin1999influence}
Peter Solin, Peter Bergin, Meroula Richardson, David~M Kaye, E~Haydn Walters,
  and Matthew~T Naughton.
\newblock Influence of pulmonary capillary wedge pressure on central apnea in
  heart failure.
\newblock \emph{Circulation}, 99\penalty0 (12):\penalty0 1574--1579, 1999.

\bibitem[Um et~al.(2017)Um, Pfister, Pichler, Endo, Lang, Hirche, Fietzek, and
  Kuli{\'c}]{um2017data}
Terry~T Um, Franz~MJ Pfister, Daniel Pichler, Satoshi Endo, Muriel Lang, Sandra
  Hirche, Urban Fietzek, and Dana Kuli{\'c}.
\newblock Data augmentation of wearable sensor data for parkinson’s disease
  monitoring using convolutional neural networks.
\newblock In \emph{Proceedings of the 19th ACM International Conference on
  Multimodal Interaction}, pages 216--220, 2017.

\bibitem[Wagner et~al.(2020)Wagner, Strodthoff, Bousseljot, Kreiseler, Lunze,
  Samek, and Schaeffter]{wagner2020ptb}
Patrick Wagner, Nils Strodthoff, Ralf-Dieter Bousseljot, Dieter Kreiseler,
  Fatima~I Lunze, Wojciech Samek, and Tobias Schaeffter.
\newblock {PTB-XL}, a large publicly available electrocardiography dataset.
\newblock \emph{Scientific data}, 7\penalty0 (1):\penalty0 1--15, 2020.

\bibitem[Wen et~al.(2020)Wen, Sun, Yang, Song, Gao, Wang, and Xu]{wen2020time}
Qingsong Wen, Liang Sun, Fan Yang, Xiaomin Song, Jingkun Gao, Xue Wang, and
  Huan Xu.
\newblock Time series data augmentation for deep learning: A survey.
\newblock \emph{arXiv preprint arXiv:2002.12478}, 2020.

\bibitem[Yancy et~al.(2013)Yancy, Jessup, Bozkurt, Butler, Casey, Drazner,
  Fonarow, Geraci, Horwich, Januzzi, et~al.]{yancy20132013}
Clyde~W Yancy, Mariell Jessup, Biykem Bozkurt, Javed Butler, Donald~E Casey,
  Mark~H Drazner, Gregg~C Fonarow, Stephen~A Geraci, Tamara Horwich, James~L
  Januzzi, et~al.
\newblock 2013 accf/aha guideline for the management of heart failure:
  executive summary: a report of the american college of cardiology
  foundation/american heart association task force on practice guidelines.
\newblock \emph{Journal of the American College of Cardiology}, 62\penalty0
  (16):\penalty0 1495--1539, 2013.

\end{thebibliography}

\clearpage
\appendix

\section{Augmentation Methods}
\label{sec:app:augs}
In this section, we provide further details on the different augmentation strategies explored (existing and TaskAug), and visualize their operation.

\subsection{Existing methods}
\begin{figure}[h]
     \centering
         \centering
         \includegraphics[width=\linewidth]{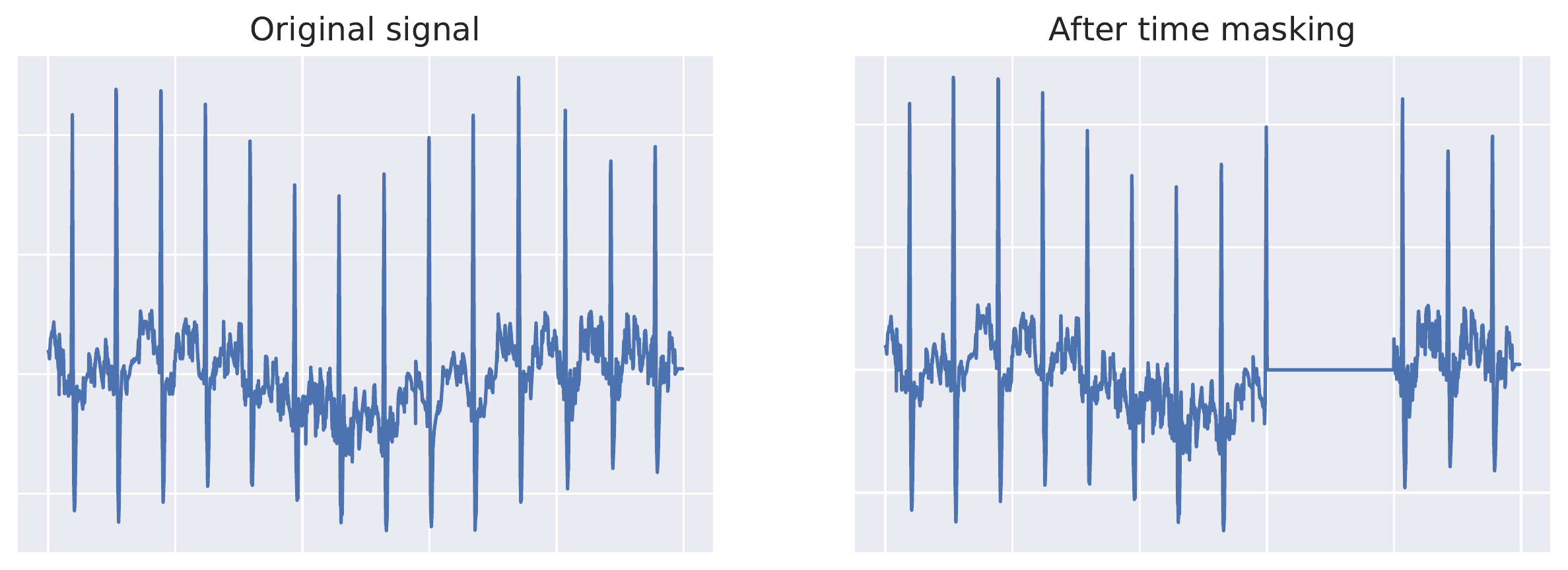}
        \caption{\small Time Masking.}
        \label{fig:timemask}
\end{figure}

\begin{figure}[h]
     \centering
         \centering
         \includegraphics[width=\linewidth]{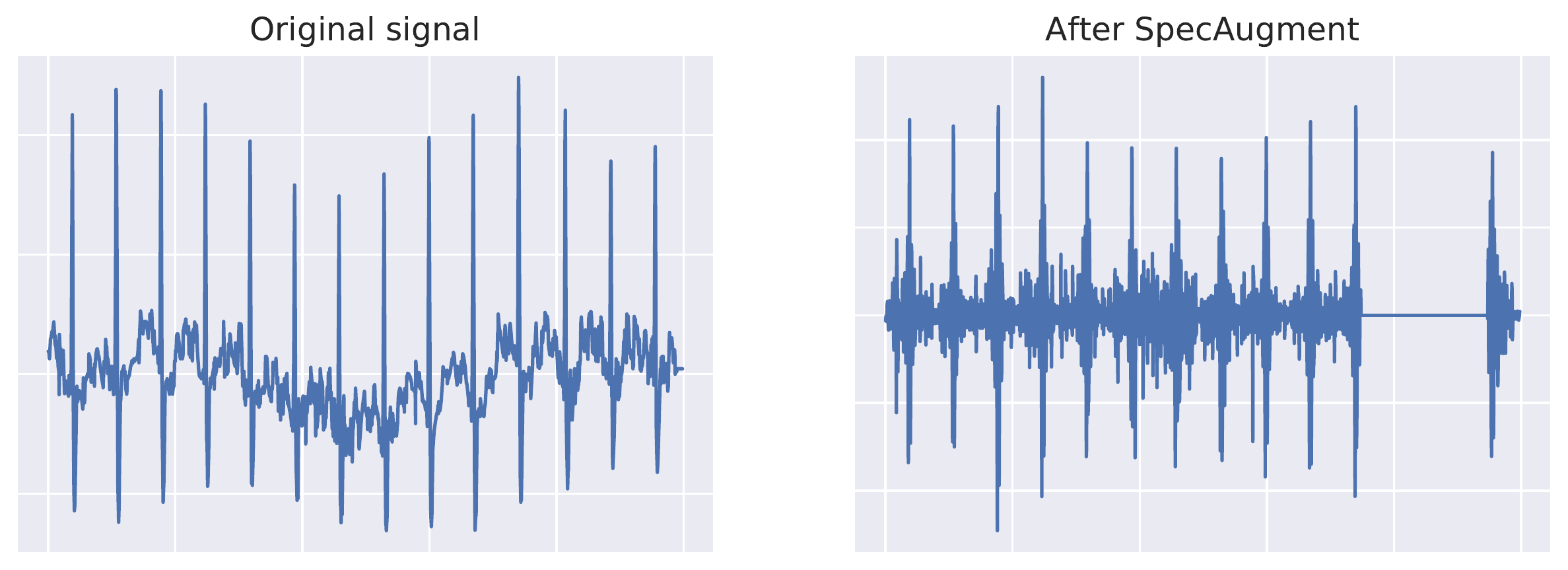}
        \caption{\small SpecAugment.}
        \label{fig:SpecAug}
\end{figure}

\begin{figure}[h]
     \centering
         \centering
         \includegraphics[width=\linewidth]{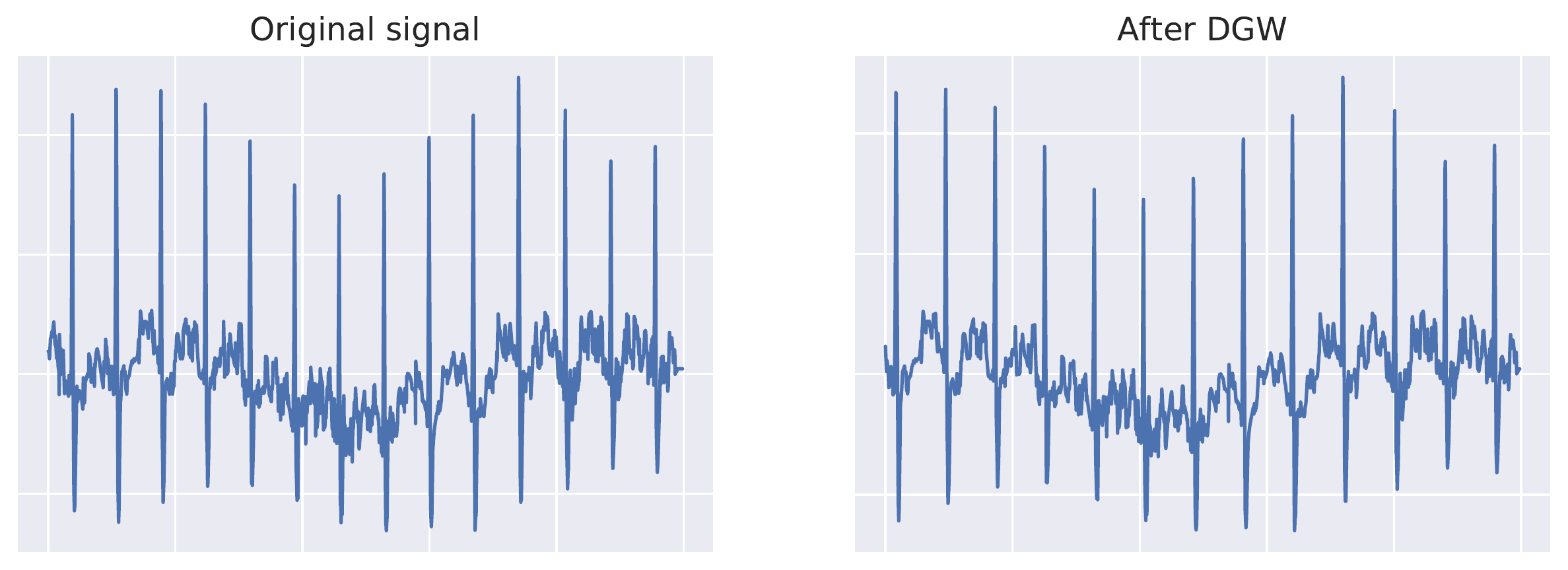}
        \caption{\small Discriminative Guided Warping (DGW).}
        \label{fig:dgw}
\end{figure}

\begin{figure}[h]
     \centering
         \centering
         \includegraphics[width=\linewidth]{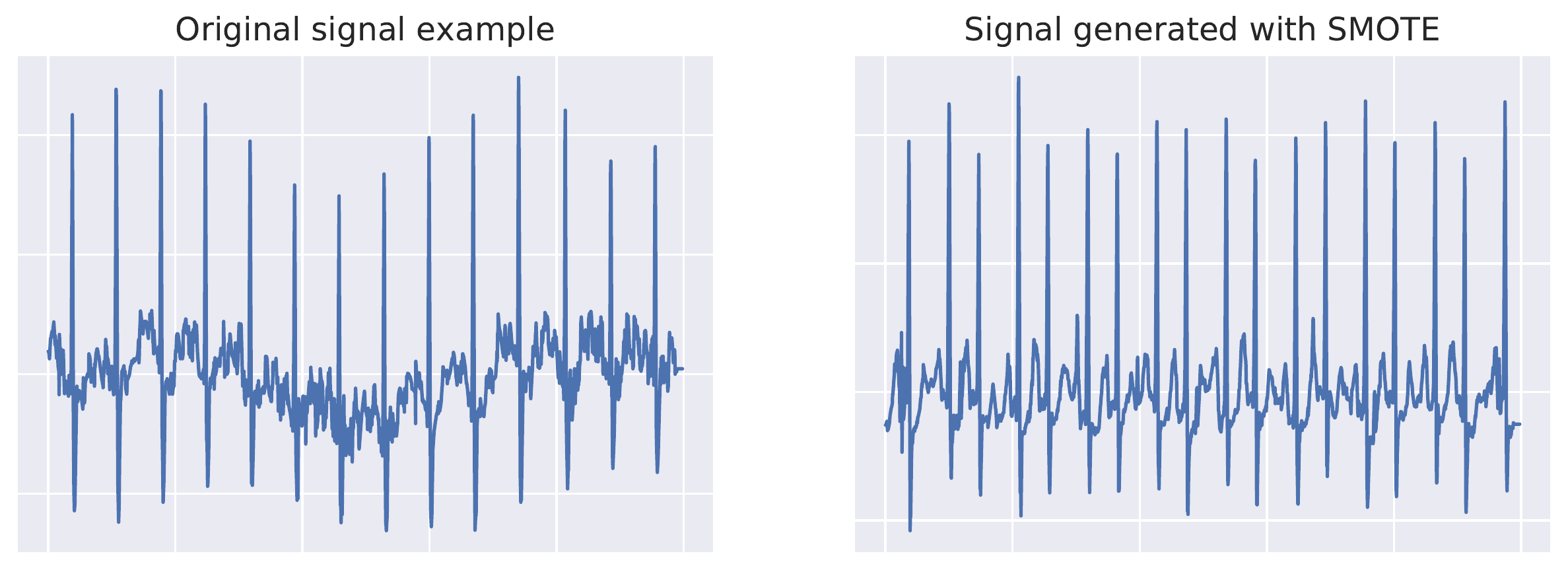}
        \caption{\small SMOTE.}
        \label{fig:smote}
\end{figure}

Figures \ref{fig:timemask}-\ref{fig:smote} present examples following augmentation using the existing methods. We show only one lead for clarity; however, these operations will be applied to each lead. 

\subsection{TaskAug}
We provide more details about TaskAug:  (1) further information about the mathematical formalism of the policy and an example of applying the different steps; (2) A more detailed description of the nested optimization algorithm used to learn TaskAug parameters, including a full algorithm; and (3) mathematical descriptions of the operations used in TaskAug in our experiments and a visualization of their effect on an ECG signal.

\subsubsection{Structure of policy}
\paragraph{Mathematical definition.} As described in Section \ref{sec:taskaug-defn}, the TaskAug policy is defined following \citet{hataya2020meta}. At each augmentation stage $k \in \{1, \ldots, K\}$ we have a set of operation selection parameters $\mathbf{ \pi}^{(k)} \in [0,1]^{M}$, where $\sum_{i} \pi_i^{(k)} = 1 \ \ \forall k$. Each vector $\pi^{(k)}$ parameterizes a categorical distribution such that each entry $\pi_i^{(k)}$ represents the probability of selecting operation $i$ at augmentation stage $k$. We obtain a reparameterizable sample from this categorical distribution (using the Gumbel-Softmax trick, \citep{jang2016categorical,maddison2016concrete}) at each stage to select the operation to use, as follows:
\begin{align}
    u &\sim \textnormal{Categorical}(\pi^{(k)}) \label{eq:app:pol1} \quad \texttt{\# Note that $u \in \mathbb{R}^M$}\\
    i &= \arg\max u  \label{eq:app:pol2} \\
    \tilde{x} &= \frac{u_i}{\texttt{stop\_grad}(u_i)} A_i(x, y; \mu_0, \mu_1) \label{eq:app:pol3}.
\end{align}

\paragraph{Why the multiplicative factor?} We use the multiplicative factor $\frac{u_i}{\texttt{stop\_grad}(u_i)}$ to allow gradient flow to the operation selection parameters $\pi$. If we just selected $i = \arg\max u$ and had no scaling in Eqn \ref{eq:app:pol3}, then there would be no gradient flow to $\pi$, since the $\arg \max$ operation is not differentiable.  

The denominator of this scaling factor is necessary because $u_i$, obtained from the reparameterized sample from the categorical distribution, is not one-hot. The resulting fraction used as the scaling factor always has magnitude 1, since $|\texttt{stop\_grad}(u_i)| = |u_i|$. When we take the gradient, we get:
$$
    \frac{\partial}{\partial \pi} \frac{u_i}{\texttt{stop\_grad}(u_i)} = \frac{1}{\texttt{stop\_grad}(u_i)} \frac{\partial{u_i}}{\partial \pi},
$$
so the $\texttt{stop\_grad}(u_i)$ acts as a scaling term.

\paragraph{Example application of TaskAug.} Suppose we have a one-stage TaskAug policy, $K=1$, our augmentation set has two operations $\mathcal{S} = \{A_1, A_2\}$ which are $A_1 =$ TimeMask$(x,y;\mu_0=0.2, \mu_1=0.1)$ and $A_2 = $ Noise$(x,y;\mu_0=2.1, \mu_1=5.3)$, and the operation selection probability vector is $\pi = [0.9, 0.1]$ (that is, we select TimeMask with probability 0.9, and noise with probability 0.1). Now consider applying TaskAug to a (data, label) pair $(x,1)$, i.e., the label is 1. We follow these steps:
\begin{enumerate}[nosep, leftmargin=*]
    \item Obtain a reparameterizable sample $u$ from Categorical($[0.9,0.1]$): let this be \mbox{$u = [0.75, 0.25]$}.
    \item Find $i = \arg\max u$; in this case, $i=1$.
    \item Select the operation $A_1$, i.e. TimeMask.
    \item Compute the masking strength based on the label. Recall this is defined as $s = y \mu_1 + (1-y) \mu_0$, so $s = 1 \times 0.1 + (1-1) \times 0.2 = 0.1$.
    \item Apply time-masking with strength $0.1$ to $x$, generating $\hat x$.
    \item Scale this by $\frac{u_1}{\texttt{stop\_grad}(u_1)}$ to generate $\tilde x$.
\end{enumerate}

\subsubsection{Parameter optimization}
As detailed in the main text, there are many learnable parameters in TaskAug, and we use gradient-based optimization to learn these jointly with the base model parameters.  Here, we provide some more details about the estimation of the gradient wrt the TaskAug parameters, and also include a full algorithm detailing the training procedure, Algorithm \ref{ALGORITHM}.

\paragraph{Estimating TaskAug parameter gradients.} Let the base model parameters after $P$ update steps be denoted as $\hat \theta(\phi)$. We update the TaskAug policy parameters to minimize the base model's validation loss $\mathcal{L}_V$, with the gradient of interest being:
$$\frac{\partial \mathcal{L}_V}{\partial \phi} = \frac{\partial \mathcal{L}_V}{\partial \hat \theta} \times \frac{\partial \hat \theta}{\partial \phi}.$$
The first term on the RHS can be found exactly using standard backpropagation. To compute the second term, we re-express it using the implicit function theorem (IFT) as in \citet{lorraine2020optimizing}. Using $\mathcal{L}_T$ to denote the training loss, the IFT allows us to re-express this second term as:
\begin{align}
\frac{\partial \hat \theta}{\partial \phi} = - \left[ \frac{\partial^2 \mathcal{L}_T}{\partial \theta \, \partial \theta^T}\right]^{-1} \times \frac{\partial^2 \mathcal{L}_T}{\partial \theta \, \partial \phi^T} \, \, \Bigr|_{\substack{\hat \theta(\phi)}}, \label{eqn:ift}
\end{align}
which is a product of an inverse Hessian and a matrix of mixed partial derivatives. Adopting the algorithm from \citet{lorraine2020optimizing}, we approximate this with a truncated Neumann series with 1 term, and implicit vector-Jacobian products.

\paragraph{Training algorithm.}  Incorporating this gradient estimator, the algorithm to jointly optimize base model parameters and TaskAug policy parameters is given in Algorithm \ref{ALGORITHM}, mirroring the approach used in \citet{raghu2021teaching}.

\begin{algorithm}
\caption{Optimizing TaskAug parameters.}
\begin{algorithmic}[1]
  \STATE Initialize base model parameters $\theta$ and TaskAug parameters $\phi$
  \FOR{$t=1, \ldots, T$}
    \STATE Compute training loss, $\mathcal{L}_T(\theta)$
    \STATE Compute $\frac{\partial \mathcal{L}_T}{\partial \theta}$
    \STATE Update $\theta \leftarrow \theta - \eta_\theta \frac{\partial \mathcal{L}_T}{\partial \theta}$
    \IF{$t \ \% \ P == 0$} 
      \STATE Set $\hat \theta = \theta$
      \STATE Compute the validation loss, $\mathcal{L}_V(\hat \theta)$
      \STATE Compute $\frac{\partial \mathcal{L}_V}{\partial \hat \theta}$
      \STATE Approximate $\frac{\partial{\hat{\theta}}}{\partial \phi}$ using Equation \ref{eqn:ift}.
      \STATE Compute the derivative $\frac{\partial \mathcal{L}_V}{\partial \phi} = \frac{\partial \mathcal{L}_V}{\partial \hat \theta} \times \frac{\partial \hat \theta}{\partial \phi}$ using the previous two steps.
      \STATE Update $\phi \leftarrow \phi - \eta_\phi \frac{\partial \mathcal{L}_V}{\partial \phi}$
    \ENDIF
  \ENDFOR
\end{algorithmic}
\label{ALGORITHM}
\end{algorithm}

\paragraph{Choice of $P$.} The value of $P$ influences how many `inner' gradient steps (to the base model) we perform before an `outer' gradient step (to the TaskAug parameters). There is a tradeoff here: if $P$ is too small, then applying the IFT to approximate $\frac{\partial \hat \theta}{\partial \phi}$ will result in a poor approximation \citep{lorraine2020optimizing}; if $P$ is too large, then updates to the policy parameters will have little effect on model parameters since the base model has already reached minimal training loss (and may start to overfit). In our experiments, we find that $P > 5$ suffered from this second problem, and $P = 1$ was sometimes unstable due to the first problem. In general, $P=1$ worked well at small sample sizes ($N=1000$), and $P=5$ worked better at $N=2500$ and $N=5000$.

\subsubsection{Augmentation operations}
\begin{figure*}[h]
     \centering
         \centering
         \includegraphics[width=\linewidth]{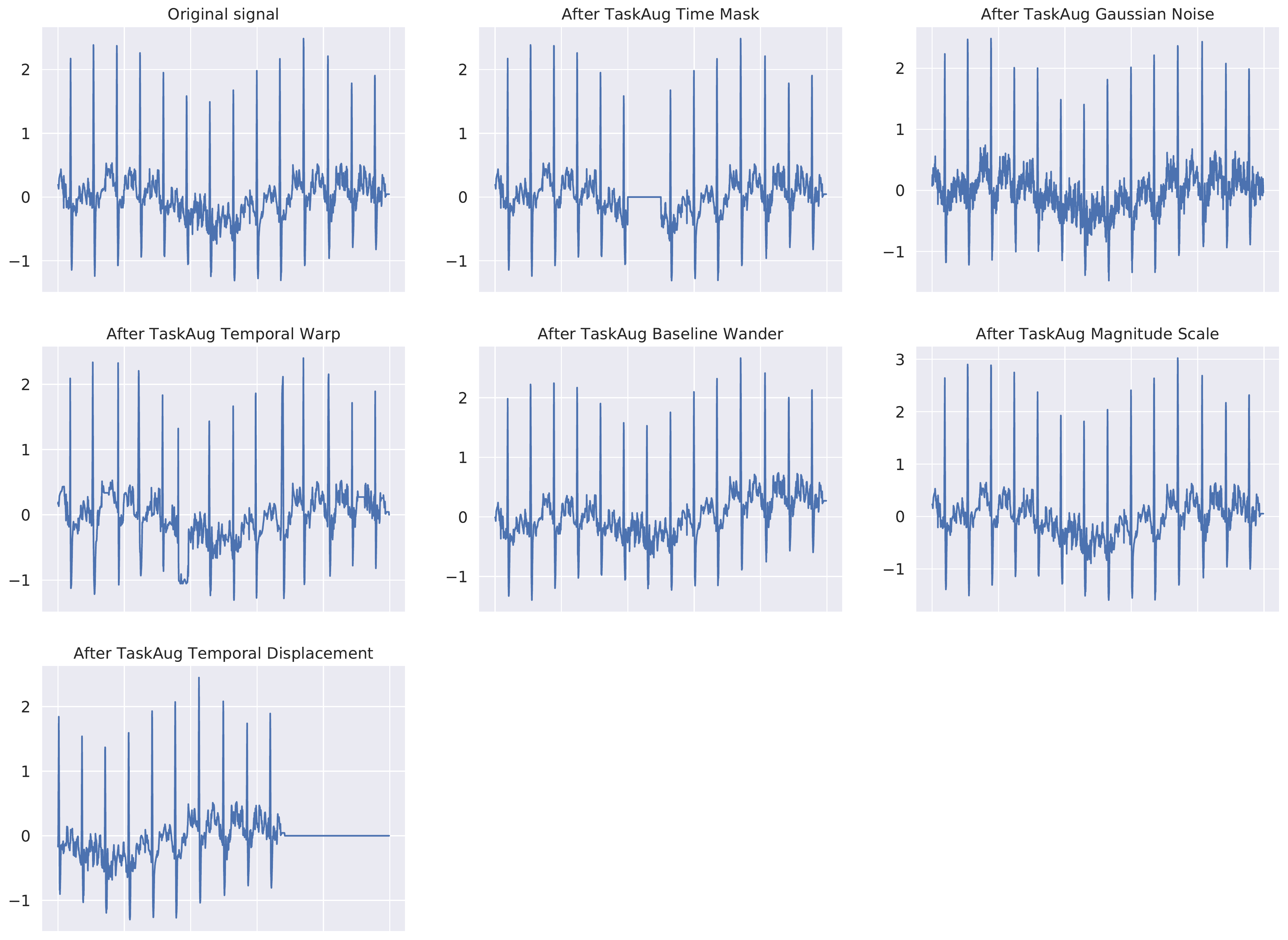}
        \caption{\small Examples of the different operations used in TaskAug.}
        \label{app:fig:taskaug}
\end{figure*}
Figure \ref{app:fig:taskaug} shows the different operations used in TaskAug. We show only one lead for clarity; however, these operations will be applied to each lead. We now provide more details on the implementation of these operations in our experiments.
\begin{itemize}[nosep, leftmargin=*]
    \item \textbf{TimeMask.} As with the existing TimeMask strategies, we randomly select a contiguous portion of the signal to set to zero. We set 10\% of the signal to zero in our implementation. This parameter is not optimized.
    \item \textbf{Gaussian Noise.} IID Gaussian noise is added to the signal. This is formed as follows. We first compute the standard deviation of each lead of the signal: let us denote this as $\sigma$. Then, the noise added to each sample of the signal is expressed as: $\epsilon = 0.25 \times \sigma \times \texttt{sigmoid}(s) \times \mathcal{N}(0,1)$, where $s$ is the learnable strength parameter, initialized to 0. The coefficient $0.25$ was found by visual inspection of some augmented examples, and observing that this allowed flexible augmentations to be generated without overwhelming the signal with noise.
    \item \textbf{Temporal warping.} The signal is warped with a random, diffeomorphic temporal transformation. To form this, we sample from a Gaussian with zero mean, and a fixed variance $100\times s^2$, where $s$ is the learnable strength parameter (initialized to 1), at each temporal location, to generate a length $T$ dimensional random velocity field. This velocity field is then integrated (following the scaling and squaring numerical integration routine used by \cite{balakrishnan2018reg,balakrishnan2019tmi}). This resulting displacement field is then smoothed with a Gaussian filter to generate the smoothed temporal displacement field. This field represents the number of samples each point in the original signal is translated in time. The field is then used to transform the signal, translating each channel in the same way (i.e., the field is the same across channels). \item \textbf{Baseline wander.}  We firstly form a wander amplitude by computing: $A = 0.25 \times \texttt{sigmoid}(s) \times \texttt{Uniform}(0,1)$, where again $s$ is a learnable strength parameter. Then, we compute the frequency and phase of the sinusoidal offset. The frequency is computed as: $f = \frac{20 \times \texttt{Uniform}(0,1) + 10}{60}$, based on the approximate number of breaths per minute for an adult. The phase is: $\phi = 2\pi \times  \texttt{Uniform}(0,1)$. Then, the sinusoidal offset is computed as: 
    $A \sin (ft + \phi)$. 
    \item \textbf{Magnitude scaling.} We scale the entire signal by a random magnitude given by \mbox{$\texttt{sigmoid}(s) \times \texttt{Uniform}(0.75,1.25)$}, where $s$ is a learnable strength parameter, initialized to 0.
    \item \textbf{Temporal displacement.} We shift the entire signal in time, padding with zeros where required.  Our implementation directly generates a displacement field (as with temporal warping) and uses the spatial transformation from \cite{balakrishnan2018reg,balakrishnan2019tmi} to transform the signal. This allows the operation to be differentiable, and for us to learn the displacement strength $s$. The displacement magnitude is a Uniform distribution on $[-100\times s^2, 100\times s^2]$, with the strength being initialized to 0.5.
\end{itemize}

\clearpage

\section{Dataset Details}
\label{sec:app:dataset}
We provide more details about the three datasets.

\subsection{Dataset A}
The labels for RVH and AFib were assigned to each example based on whether relevant diagnostic statements were present in either a clinician's read of the ECG, or a machine read of the ECG. 

For RVH, there were six diagnostic statements that led to a positive label being assigned: ``right ventricular hypertrophy", ``biventricular hypertrophy", ``combined ventricular hypertrophy", ``right ventricular enlargement", ``rightventricular hypertrophy", ``biventriclar hypertrophy".

For AFib, there were nine such statements: ``atrial fibrillation with rapid ventricular response", ``atrial fibrillation with moderate ventricular response", ``fibrillation/flutter", ``atrial fibrillation with controlled ventricular response", ``afib", ``atrial fib", ``afibrillation", ``atrial fibrillation", ``atrialfibrillation".


\paragraph{Preprocessing.} ECGs were sampled at 250 Hz for 10 seconds, resulting in a $2500 \times 12$ tensor for all 12 leads, per-ECG. We normalized the signals by dividing by 1000. Other forms of normalization for this dataset (e.g., z-scoring) resulted in some abnormally large/small values.

\subsection{Dataset B}
The four labels are obtained by aggregating relevant sets of diagnostic statements -- we refer the reader to the PTB-XL paper \citep{wagner2020ptb} for further details. Of relevance here is that certain labels, such as MI, contain a small number of distinct diagnostic statements (3), potentially suggesting why many augmentation strategies can help -- it is a fine-grained task. Others (such as CD) are much broader, covering many more diagnostic statements. 

\paragraph{Preprocessing.} ECGs in the dataset are sampled at 500 Hz for 10 seconds; we downsample these by a factor of 2 for consistency with Dataset A and C, resulting in a $2500 \times 12$ tensor for all 12 leads, per-ECG. Normalization involved z-scoring, following the code provided with the dataset.

\subsection{Dataset C}
The hemodynamics prediction cohort consists of patients who had an ECG and right heart catheterization procedure on the same day. The catheterization procedure measures hemodynamics variables including the pulmonary capillary wedge pressure (PCWP) and cardiac output (CO), and these are used to form the prediction targets.  We consider inferring abnormally low Cardiac Output (less than 2.5 L/min), and abnormally high Pulmonary Capillary Wedge Pressure (greater than 20 mmHg).

\paragraph{Preprocessing.} ECGs were sampled at 250 Hz for 10 seconds, resulting in a $2500 \times 12$ tensor for all 12 leads, per-ECG. We normalized the signals by dividing by 1000. Other forms of normalization for this dataset (e.g., z-scoring) resulted in some abnormally large/small values, so we opted for the division-based normalization.

\clearpage

\section{Experiments}
\label{sec:app:expts}
In this section, we provide further experimental details. We first provide implementation details, and then outline additional experimental results including: Results for AUPRC in the low-sample ($N=1000$) regime,  performance on Datasets A and B in the high-sample regime, performance on Dataset B in an additional low sample regime ($N=500$ data points), interpretation of the TaskAug policy for RVH, a study of the impact of optimizing policy parameters across different sample size regimes, and a study of the impact of class-specific magnitudes across different sample size regimes. 

\subsection{Implementation details}
\paragraph{Network architecture.} In all experiments, we use a 1D CNN based on a ResNet-18 \citep{he2016deep} architecture. This model has convolutions with a kernel size of 15, and stride 2 (informed by the temporal window we want the convolutions to operate over). The blocks in the ResNet architecture have convolutional layers with 32, 64, 128, and 256 channels respectively. The output after the final block is average pooled in the temporal dimension, and then a linear layer is applied to predict the probability of the positive class. 

\paragraph{Optimization settings.} As discussed, we used Adam with a learning rate of 1e-3 for all methods, given that this resulted in stable training across all settings. When optimizing the TaskAug policy parameters, we used RMSprop with a learning rate of 1e-2, following \citet{lorraine2020optimizing}.

\paragraph{Computational information.} All models and training were implemented in PyTorch and run on a single NVIDIA V100 GPU. 

\subsection{Additional results}
\paragraph{AUPRC results at 1000 samples.} As discussed in Section \ref{sec:results}, the improvements in AUROC are not always statistically significant. Given that some of the labels are very low prevalence (RVH: 1\%, AFib: 5\%, low CO: 4\%), we evaluate the AUPRC in the low-sample regime, which provides additional information about model performance.  Results are shown in Tables \ref{app:tab:muse-1000-auprc}, \ref{app:tab:ptbxl-1000-auprc}, and \ref{app:tab:hemo-auprc}. We observe that for the low prevalence RVH, AFib, and Low CO tasks, TaskAug obtains statistically significant improvements in performance. On Dataset A tasks (RVH and AFib), it is the only method to do so.

\begin{table}[htbp]
\centering
\begin{tabular}{@{}lll@{}}
\toprule
    &         \multicolumn{1}{c}{RVH}                     & \multicolumn{1}{c}{AFib}                     \\ \midrule
NoAugs      & 7.4 $\pm$ 1.3                               & \underline{21.2 $\pm$ 2.0}                               \\
TaskAug      & \textbf{10.8 $\pm$ 0.8}$^*$                     & \textbf{27.3 $\pm$ 1.8}$^*$      \\
SMOTE        & 9.7 $\pm$ 1.2                               & 21.0 $\pm$ 2.2                                                          \\
DGW          & 7.1 $\pm$ 0.9                               & 19.4 $\pm$ 2.3                           \\
SpecAug  & \underline{10.6 $\pm$ 1.2}                  & 21.1 $\pm$ 2.0                     \\
TimeMask & 10.1 $\pm$ 1.5                              & 20.3 $\pm$ 2.3                            \\ \bottomrule
\end{tabular}
\caption{\small Mean and standard error of AUPRC for various data augmentation strategies when detecting cardiac abnormalities on Dataset A. We consider a low-sample regime with a development set of 1000 data points. The best-performing method is bolded, and the second best is underlined, and $^*$ indicates statistically significant improvement at the $p<0.05$ level. TaskAug is the only method to obtain significant improvements in performance on both tasks.}
\label{app:tab:muse-1000-auprc}
\end{table}

\begin{table*}[]
\centering
\begin{tabular}{@{}lllll@{}}
\toprule
\textbf{}    & \multicolumn{1}{c}{MI} & \multicolumn{1}{c}{HYP} & \multicolumn{1}{c}{STTC} & \multicolumn{1}{c}{CD} \\ \midrule
NoAugs      & 59.2$\pm$2.1                & 53.1$\pm$1.7               & 66.9$\pm$2.5                & \textbf{67.3$\pm$1.1}      \\
TaskAug      & \textbf{63.1$\pm$1.7}       & \textbf{55.2$\pm$0.9}      & 68.7$\pm$1.3                & 66.8$\pm$1.2               \\
SMOTE        & {\ul 62.0$\pm$1.6}          & 41.2$\pm$2.9               & 65.9$\pm$1.0                & 62.7$\pm$1.1               \\
DGW          & 61.1$\pm$1.2                & 53.9$\pm$1.6               & 67.9$\pm$1.1                & 64.7$\pm$2.6               \\
SpecAug  & 61.7$\pm$1.6                & {\ul 54.5$\pm$1.5}         & {\ul 68.8$\pm$1.5}          & 65.8$\pm$1.4               \\
TimeMask & 60.3$\pm$1.3                & 52.8$\pm$1.8               & \textbf{68.8$\pm$1.2}       & \textbf{70.1$\pm$1.3}      \\ \bottomrule
\end{tabular}
\caption{\small Mean and standard error of AUPRC for various data augmentation strategies on detecting cardiac abnormalities on Dataset B. We consider a low-sample regime with a development set of 1000 data points. The best-performing method is bolded, and the second best is underlined, and $^*$ indicates statistically significant improvement at the $p<0.05$ level. }
\label{app:tab:ptbxl-1000-auprc}
\end{table*}

\begin{table*}[]
\centering
\begin{tabular}{@{}llll@{}}
\toprule
\textbf{}    & {Low CO}      & \begin{tabular}[c]{@{}l@{}}High PCWP: \\ $N=1000$ \end{tabular}      & \begin{tabular}[c]{@{}l@{}}High PCWP: \\ All Data \end{tabular}    \\ \midrule
NoAugs      & 7.2 $\pm$ 0.4                              & \underline{42.5 $\pm$ 0.8} & 49.7 $\pm$ 0.8                              \\
TaskAug      & \textbf{8.8 $\pm$ 0.6}$^*$    & \textbf{43.5 $\pm$ 0.9}    & \textbf{50.8 $\pm$ 0.8}    \\
SMOTE        & \textbf{8.8 $\pm$ 0.6}$^*$    & 41.9 $\pm$ 0.7                              & 46.9 $\pm$ 0.7                              \\
DGW          & \underline{8.1 $\pm$ 0.7} & 41.2 $\pm$ 0.7                              & 49.7 $\pm$ 1.0                              \\
SpecAug  & 7.8 $\pm$ 0.4                              & 42.3 $\pm$ 1.1                              & \underline{50.3 $\pm$ 0.8} \\
TimeMask & 8.0 $\pm$ 0.5                              & 42.4 $\pm$ 0.7                              & 50.1 $\pm$ 0.9                              \\ \bottomrule
\end{tabular}
\caption{\small Mean and standard error of AUPRC for various data augmentation strategies for the hemodynamics inference task in Dataset C. We consider a low-sample regime with a development set of 1000 data points. The best-performing method is bolded, and the second best is underlined, and $^*$ indicates statistically significant improvement at the $p<0.05$ level. TaskAug is the one of only two methods to obtain significant improvements in performance on the low CO detection task.}
\label{app:tab:hemo-auprc}
\end{table*}

\clearpage
\paragraph{Results at higher sample regimes.} Tables \ref{app:tab:muse2500-auroc}-\ref{app:tab:ptbxl-5000-auroc} show AUROC for the different augmentation methods on the tasks from Datasets A and B. 
We observe that augmentations are less effective at higher samples. Particularly when the development set sizes are 2500 and 5000 datapoints, we observe that the improvement with using augmentations (over the NoAugs baseline) with any of the methods is quite small, and nearly always less than 1\% AUROC. This suggests that in general, augmentations are less useful at these higher data regimes.

\begin{table}[htbp]
\centering
\begin{tabular}{@{}lll@{}}
\toprule
    & \multicolumn{1}{c}{RVH} & \multicolumn{1}{c}{AFib}  \\ \midrule
NoAugs      & {\ul 86.1$\pm$0.9}         & 89.0 $\pm$ 0.4                           \\
TaskAug      & \textbf{86.9$\pm$0.9}      & {\ul 89.1 $\pm$ 0.4}                  \\
SMOTE        & 85.5$\pm$1.3               & 89.1 $\pm$ 0.5                             \\
DGW          & 84.8$\pm$1.3               & 88.4 $\pm$ 0.5                     \\
SpecAug  & 83.3$\pm$1.8               & \textbf{89.1 $\pm$ 0.3}                     \\
TimeMask & 85.8$\pm$1.1               & 88.2 $\pm$ 0.4                             \\ \bottomrule
\end{tabular}
\caption{\small Mean and standard error of AUROC for augmentation methods on Dataset A tasks with a development set of 2500 data points. The best performing method is bolded, and the second best is underlined.}
\label{app:tab:muse2500-auroc}
\end{table}

\begin{table}[htbp]
\centering
\begin{tabular}{@{}llll@{}}
\toprule
\textbf{}    & \multicolumn{1}{c}{RVH} & \multicolumn{1}{c}{AFib}  \\ \midrule
NoAugs      & {\ul 90.6$\pm$0.6}         & 92.6$\pm$0.2                            \\
TaskAug      & 90.6$\pm$0.4               & \textbf{92.8$\pm$0.1}          \\
SMOTE        & 89.8$\pm$0.6               & 92.6$\pm$0.2                            \\
DGW          & \textbf{90.8$\pm$0.5}      & 92.5$\pm$0.2                        \\
SpecAug  & 90.5$\pm$0.8               & {\ul 92.7$\pm$0.1}                      \\
TimeMask & 89.4$\pm$0.7               & 92.6$\pm$0.2                         \\ \bottomrule
\end{tabular}
\caption{\small Mean and standard error of AUROC for augmentation methods on Dataset A tasks with a development set of 5000 data points. The best performing method is bolded, and the second best is underlined.}
\label{app:tab:muse5000-auroc}
\end{table}


\begin{table*}[]

\centering
\begin{tabular}{@{}lllll@{}}
\toprule
\textbf{}    & \multicolumn{1}{c}{MI} & \multicolumn{1}{c}{HYP} & \multicolumn{1}{c}{STTC} & \multicolumn{1}{c}{CD} \\ \midrule
NoAugs      & 84.5$\pm$0.5                & {\ul 86.4$\pm$0.4}         & 89.7$\pm$0.3                & 85.8$\pm$0.3               \\
TaskAug      & \textbf{86.1$\pm$0.5}       & 86.2$\pm$0.4               & 89.7$\pm$0.3                & 86.6$\pm$0.4               \\
SMOTE        & 84.7$\pm$0.7                & 81.9$\pm$1.3               & 88.7$\pm$0.4                & 85.5$\pm$0.6               \\
DGW          & 84.1$\pm$0.5                & 85.9$\pm$0.6               & 89.5$\pm$0.3                & 86.2$\pm$0.3               \\
SpecAug  & 84.6$\pm$0.8                & 86.2$\pm$0.6               & \textbf{90.2$\pm$0.3}       & {\ul 86.8$\pm$0.6}         \\
TimeMask & {\ul 85.7$\pm$0.4}          & \textbf{86.6$\pm$0.3}      & {\ul 90.1$\pm$0.1}          & \textbf{87.0$\pm$0.7}      \\ \bottomrule
\end{tabular}
\caption{\small Mean and standard error of AUROC for augmentation methods on Dataset B tasks with a development set of 2500 data points. The best performing method is bolded, and the second best is underlined.}
\label{app:tab:ptbxl-2500-auroc}
\end{table*}

\begin{table*}[]

\centering
\begin{tabular}{@{}lllll@{}}
\toprule
\textbf{}    & \multicolumn{1}{c}{MI} & \multicolumn{1}{c}{HYP} & \multicolumn{1}{c}{STTC} & \multicolumn{1}{c}{CD} \\ \midrule
NoAugs      & {\ul 89.4$\pm$0.3}          & 88.2$\pm$0.2               & 91.0$\pm$0.3                & 89.3$\pm$0.4               \\
TaskAug      & {\ul 89.4$\pm$0.3}          & 88.3$\pm$0.2               & \textbf{91.6$\pm$0.2}       & \textbf{90.0$\pm$0.2}      \\
SMOTE        & 86.6$\pm$0.7                & 86.7$\pm$0.4      & 90.6$\pm$0.3                & 88.0$\pm$0.3               \\
DGW          & 88.6$\pm$0.3                & 88.0$\pm$0.2               & {\ul 91.3$\pm$0.1}          & 89.3$\pm$0.2               \\
SpecAug  & \textbf{89.5$\pm$0.2}       & {\ul 88.4$\pm$0.4}               & \textbf{91.6$\pm$0.2}       & {\ul 89.9$\pm$0.2}         \\
TimeMask & 89.3$\pm$0.3                & {\bf 88.6$\pm$0.2}         & \textbf{91.6$\pm$0.2}       & 89.8$\pm$0.2               \\ \bottomrule
\end{tabular}
\caption{\small Mean and standard error of AUROC for augmentation methods on Dataset B tasks with a development set of 5000 data points. The best performing method is bolded, and the second best is underlined.}
\label{app:tab:ptbxl-5000-auroc}
\end{table*}

\clearpage
\paragraph{Results on Dataset B at $N=500$.} Table \ref{app:tab:ptbxl-500} shows AUROC for the different augmentation methods in an additional low sample regime, with $N=500$. We see that the maximum improvement over the NoAugs baseline by any augmentation strategy is greater in this regime than it was at $N=1000$ (see Table~\ref{tab:arrhythmia-all}). Given that the prevalence of these tasks is relatively high, we see more significant performance improvements in the $N=500$ regime. 

\begin{table*}[]
\centering
\begin{tabular}{@{}lllll@{}}
\toprule
\textbf{}    & \multicolumn{1}{c}{MI}                    & \multicolumn{1}{c}{HYP}                     & \multicolumn{1}{c}{STTC}                    & \multicolumn{1}{c}{CD}                     \\ \midrule
NoAugs      & 74.4 $\pm$ 0.9                              & \textbf{81.9 $\pm$ 0.8}    & 85.2 $\pm$ 0.5                              & 78.9 $\pm$ 1.2                              \\
TaskAug      & \textbf{78.4 $\pm$ 0.5 }   & \underline{81.5 $\pm$ 1.2} & 86.2 $\pm$ 0.4                              & \textbf{80.7 $\pm$ 0.6}    \\
SMOTE        & 75.7 $\pm$ 1.2                              & 79.2 $\pm$ 1.5                              & 85.5 $\pm$ 0.3                              & 78.6 $\pm$ 1.5                              \\
DGW          & \underline{78.2 $\pm$ 0.6} & 78.7 $\pm$ 1.2                              & 82.0 $\pm$ 1.3                              & 79.0 $\pm$ 0.9                              \\
SpecAug  & 77.8 $\pm$ 0.7                              & 81.0 $\pm$ 0.6                              & \underline{86.3 $\pm$ 0.4} & 79.3 $\pm$ 1.1                              \\
TimeMask & 77.8 $\pm$ 1.0                              & 80.9 $\pm$ 1.3                              & \textbf{86.6 $\pm$ 0.5}    & \underline{80.3 $\pm$ 0.8} \\ \bottomrule
\end{tabular}
\caption{\small Mean and standard error of AUROC for augmentation methods on Dataset B tasks with a development set of 500 data points. The best performing method is bolded, and the second best is underlined.}
\label{app:tab:ptbxl-500}
\end{table*}

\clearpage
\begin{figure*}[htbp]
\floatconts
{fig:rvh-policy}
{\caption{\small \textbf{TaskAug policy for detecting Right Ventricular Hypertrophy.} The learned TaskAug policy: probability of selecting each transformation in both augmentation stages and the optimized displacement strengths in the first stage. We show the mean/standard error of the learned parameter values over 15 runs.
Temporal operations (masking and displacement) have high probability of selection in Stage 1, which is sensible since these operations are likely to be label preserving (RVH is typically detected based on relative magnitudes of portions of beats in the ECG). We see that both positive and negative classes have similar optimized displacement augmentation strengths -- we do not expect displacement to impact the class label differently for the two classes, so this is sensible.}}
{%
     \subfigure[\small Operation selection probabilities]{\label{fig:rvh-policy-probs}%
        \includegraphics[width=0.65\linewidth]{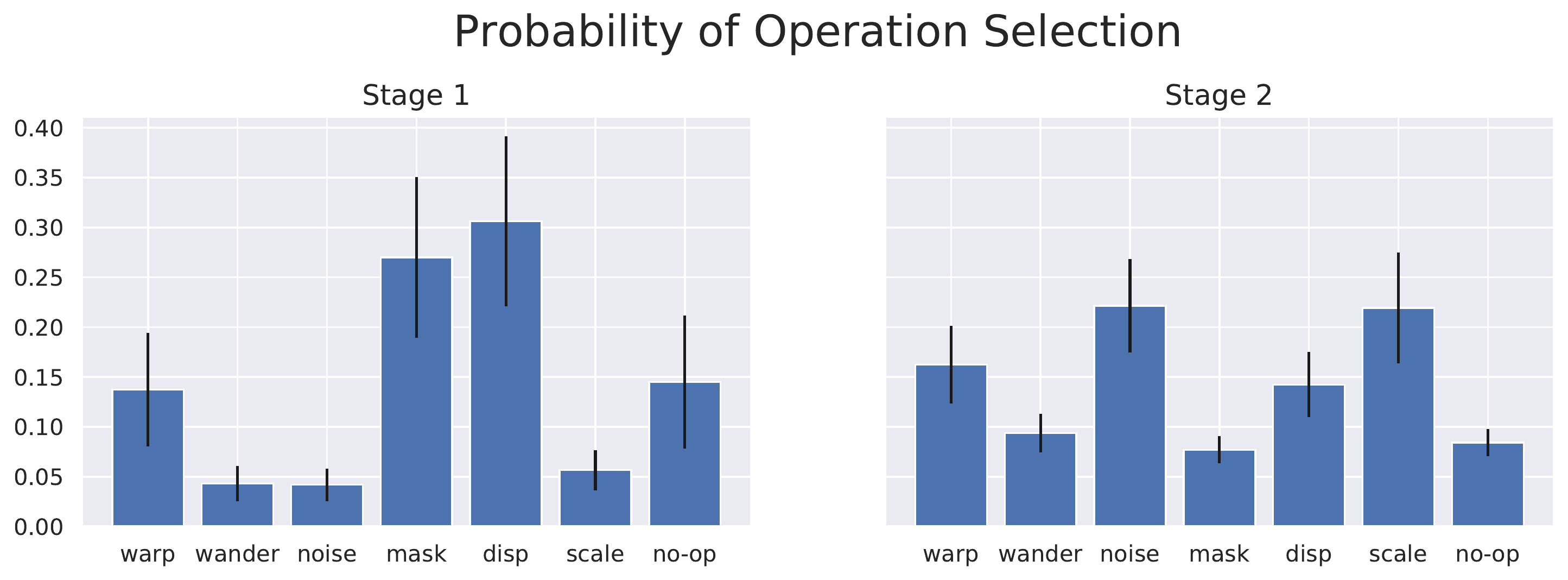}}%
        \qquad
     \subfigure[\small Displacement strengths]{\label{fig:rvh-policy-mags}%
        \includegraphics[width=0.29\linewidth]{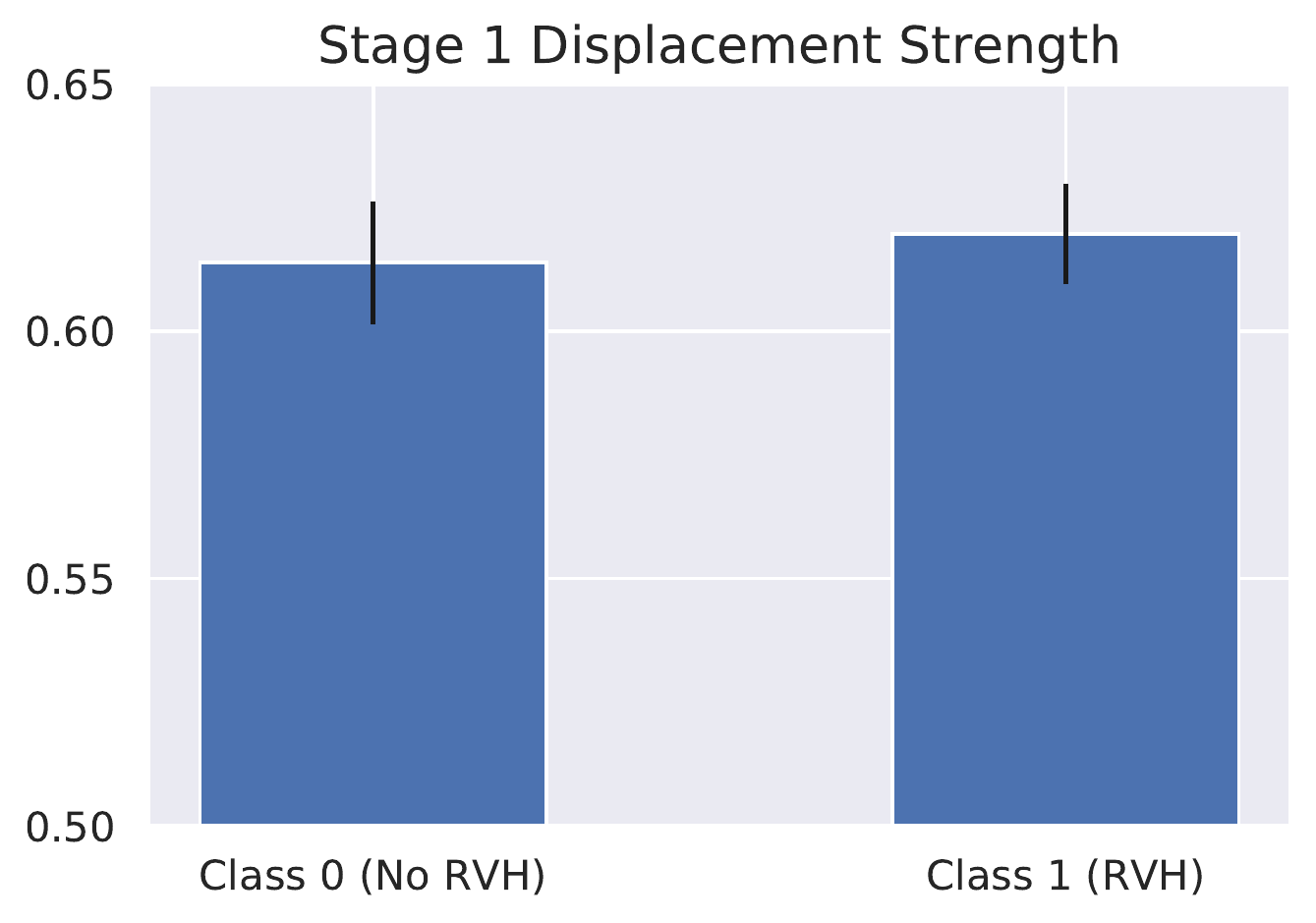}}
}
\end{figure*}
\paragraph{Interpreting the RVH policy.} We visualize the TaskAug policy for RVH in Figure~\ref{fig:rvh-policy}. We observe high probability assigned to selecting two temporal operations in stage 1, namely masking and displacement. Relative magnitudes of different portions of the ECG affect the RVH label, so temporal operations having higher probability of selection is sensible since they are more likely to be label preserving than operations that change the relative magnitudes of different parts of the ECG. We examine the learned strengths for the displacement operation in Stage 1, Figure~\ref{fig:rvh-policy-mags}, and we see that there is little differentiation on a per-class basis. This is sensible, since we do not expect displacement of the signal in time to affect the RVH label for differently for the positive and negative classes.

\begin{figure*}[]
     \centering
         \centering
         \includegraphics[width=\linewidth]{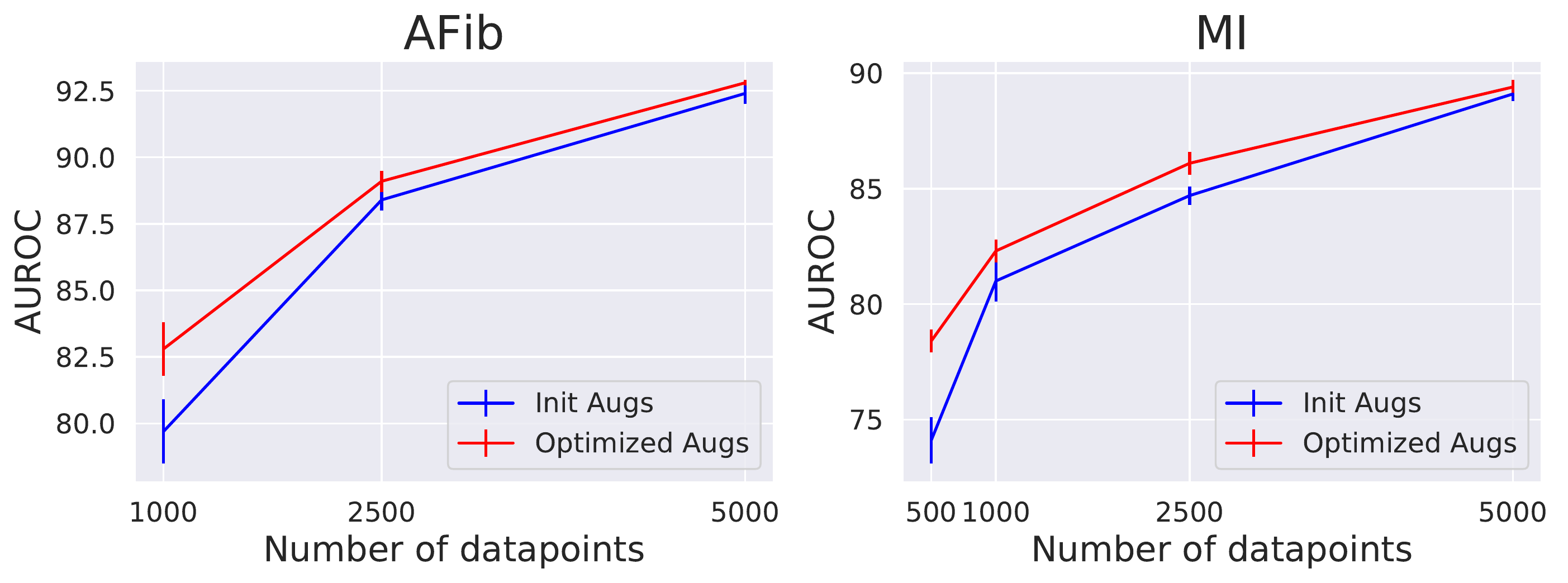}
        \caption{\small \textbf{Studying performance when we do not optimize the policy parameters in TaskAug.} We show the mean/standard error of AUROC over 15 runs for AFib and over 5 runs for MI. We see that optimizing the policy parameters results in noticeable improvements in performance over keeping the policy parameters at their initial values (InitAugs). However, the impact of optimizing the parameters is reduced at larger dataset sizes, possibly due to the fact that augmentations are inherently less useful at higher sample regimes.}
        \label{fig:ablation-app}
\end{figure*}

\clearpage
\paragraph{Further study on the impact of optimizing augmentations.} As shown in the main text, Figure \ref{fig:ablation}, optimizing the policy parameters improves performance over keeping them fixed at their initial values. In Figure \ref{fig:ablation-app}, we study this effect across different dataset sizes and find that the optimization has the most impact in the low sample regime, but still results in improvements even at higher samples. This could be due to the fact that at higher samples, augmentations boost performance less in general, so the specific parameter settings in TaskAug also have less impact.

\begin{figure*}[]
     \centering
         \centering
         \includegraphics[width=\linewidth]{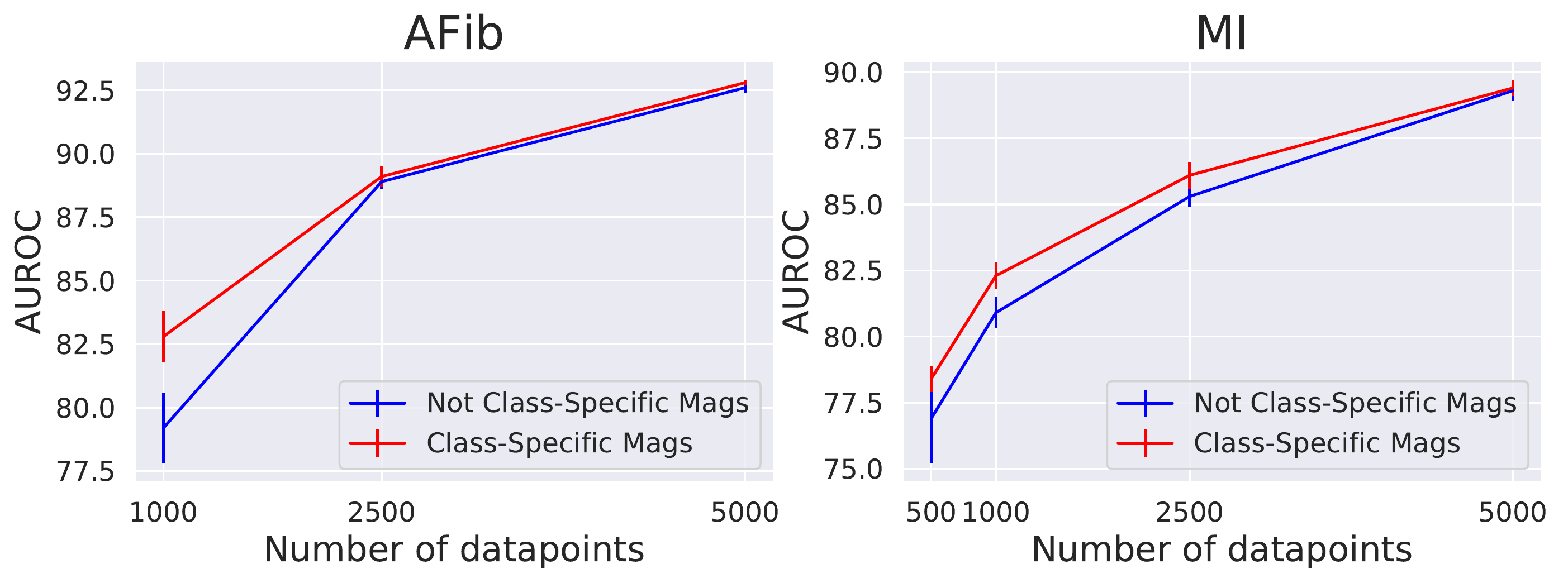}
        \caption{\small \textbf{Studying performance when we do not have class-specific magnitude parameters in TaskAug.} We show the mean/standard error of AUROC over 15 runs for AFib and over 5 runs for MI. Class-specific magnitude parameters improve performance most in the low sample regime. At higher samples, this impact is reduced, possibly due to the fact that augmentations are inherently less useful at higher sample regimes.}
        \label{fig:ablation-app-classparams}
\end{figure*}

\paragraph{Further study on the impact of class-specific magnitudes.} As shown in the main text, Figure \ref{fig:ablation2}, optimizing class-specific magnitudes improves over learning one magnitude parameter for each class. Figure \ref{fig:ablation-app-classparams} studies this effect across different dataset sizes and we see that the class-specific parameters improve performance at all dataset sizes, but the improvement is most clearly seen at low samples. Similarly with the optimization of augmentation parameters, this could be due to the fact that at higher samples, augmentations boost performance less in general, so the class-specific parameterization in TaskAug has less impact.

\end{document}